\documentclass{article}

\usepackage{PRIMEarxiv,cite}

\usepackage[utf8]{inputenc} 
\usepackage[T1]{fontenc}    
\usepackage{url}            
\usepackage{booktabs}       
\usepackage{amsfonts}       
\usepackage{nicefrac}       
\usepackage{microtype}      
\usepackage{lipsum}
\usepackage{amsmath}
\usepackage{amsthm}
\usepackage{amssymb}   

\usepackage{algorithm}
\usepackage{algpseudocode}

\usepackage{siunitx }
\usepackage{bigints}
\usepackage{bm}
\usepackage{diagbox}
\usepackage{makecell}
\usepackage{multirow}
\usepackage{caption}
\usepackage{subcaption}
\usepackage{graphicx}

\usepackage{xcolor}
\usepackage{mathtools}
\usepackage{enumitem}

\usepackage{hyperref}
\hypersetup{
    colorlinks=true,
    citecolor=black,
    linkcolor=black,
    filecolor=black,
    urlcolor=black,
}

\NewDocumentCommand\pirateflag{}{
    \includegraphics[scale=0.1]{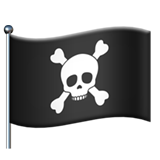}
}

\usepackage{pifont}

\newcommand{\R}{\mathbb{R}}

\def \mc {\mathcal}
\newcommand{\norm}[1]{{\left\lVert #1 \right\lVert}}
\def \rd {{\rm d}}
\def \h {\hat}
\def \p {\partial}
\def \bb {\mathbf}

\newtheorem{theorem}{Theorem}[section]

\newtheorem{corollary}[theorem]{Corollary}

\newtheorem{proposition}{Proposition}

\newtheorem{claim}{Claim}

\numberwithin{equation}{section}

\title{\pirateflag PirateNets: Physics-informed Deep Learning with Residual Adaptive Networks
}

\author{
  Sifan Wang \\
Graduate Group in Applied Mathematics \\
  and Computational Science \\
  University of Pennsylvania\\
  Philadelphia, PA 19104 \\
  \texttt { sifanw@sas.upenn.edu} \\
  \And
   Bowen Li \\
Department of Mathematics \\
  Duke University\\
  Durham, NC 27708 \\
  \texttt {bowen.li200@duke.edu} \\
  \And
    Yuhan Chen \\
Department of Electrical \\
and Computer Engineering \\
  North Carolina State University\\
  Raleigh, NC 27695 \\
  \texttt {ychen239@ncsu.edu} \\
  \And
  Paris Perdikaris \\
  Department of Mechanical Engineering \\
  and Applied Mechanics\\
  University of Pennsylvania\\
  Philadelphia, PA 19104 \\
  \texttt{pgp@seas.upenn.edu} 
}

\begin{document}
\maketitle

\begin{abstract}
While physics-informed neural networks (PINNs) have become a  popular deep learning framework for tackling forward and inverse problems governed by partial differential equations (PDEs), their performance is known to degrade when larger and deeper neural network architectures are employed. Our study identifies that the root of this counter-intuitive behavior lies in the use of multi-layer perceptron (MLP) architectures with non-suitable initialization schemes, which result in poor trainablity for the network derivatives, and ultimately lead to an unstable minimization of the PDE residual loss. To address this, we introduce Physics-Informed Residual Adaptive Networks (PirateNets), a novel architecture that is designed to facilitate stable and efficient training of deep PINN models. PirateNets leverage a novel adaptive residual connection, which allows the networks to be initialized as shallow networks that progressively deepen during training. We also show that the proposed initialization scheme allows us to encode appropriate inductive biases corresponding to a given PDE system into the network architecture. We provide comprehensive empirical evidence showing that PirateNets are easier to optimize and can gain accuracy from considerably increased depth, ultimately achieving state-of-the-art results across various benchmarks. All code and data accompanying this manuscript will be made publicly available at \url{https://github.com/PredictiveIntelligenceLab/jaxpi}.

\end{abstract}

\section{Introduction}







Machine learning (ML) is making a continual impact on the fields of science and engineering, providing advanced tools for analyzing complex data, uncovering nonlinear relationships, and developing predictive models. Notable examples include AlphaFold \cite{jumper2021highly} in protein structure prediction, Deep Potentials \cite{zhang2018deep} for large-scale molecular dynamics, and GraphCast \cite{lam2022graphcast} for medium-range weather forecasting.
The integration of physical laws and constraints within machine learning models has given rise to physics-informed machine learning (PIML). This emerging field opens new frontiers for traditional scientific research and contributes to addressing persistent challenges in machine learning, such as robustness, interpretability, and generalization.

The fundamental question that PIML aims to address is how to incorporate physical prior knowledge into machine learning models. This can be achieved by modifying key components of the machine learning pipeline, which includes data processing, model architecture, loss functions,  optimization algorithms, and fine-tuning and inference.
For example, neural operators \cite{lu2021learning, li2020fourier}  are designed to encode physics from training data that reflects underlying physical laws. 
Another powerful approach is to tweak network architectures to strictly enforce physical constraints, thereby enhancing the generalizability of the model. One method involves embedding general physical principles such as symmetries, invariances and equivariances into ML models. As highlighted by Cohen {\em et al.} \cite{cohen2019gauge} and Maron {\em et al.} \cite{maron2019universality, maron2018invariant}, this strategy leads to simpler models that require less data while achieving higher prediction accuracy. Alternatively, embedding specific physical laws can be equally effective. This is demonstrated by Mohan {\em et al.}\cite{mohan2020embedding}, who ensured continuity and mass conservation in neural networks for coarse-graining three-dimensional turbulence by integrating the curl operator. Similarly, Meng {\em et al.}\cite{darbon2020overcoming} illustrates that certain neural network architectures are inherently aligned with the physics of specific Hamilton-Jacobi (HJ) partial differential equations (PDEs).

One of the most popular and flexible methods for embedding physical principles in machine learning is through the formulation of tailored loss functions. These loss functions serve as soft constraints that bias ML models towards respecting the underlying physics during training, giving rise to the emergence of physics-informed neural networks (PINNs). Thanks to their flexibility and ease of implementation, PINNs have been extensively used to solve forward and inverse problems involving PDEs by seamlessly integrating noisy experimental data and physical laws into the learning process. In recent years, PINNs have yielded a series of promising results across various domains in computational science, including applications in fluid mechanics \cite{raissi2020hidden,sun2020surrogate, mathews2021uncovering}, bio-engineering    \cite{sahli2020physics, kissas2020machine}, and material science \cite{fang2019deep,chen2020physics, zhang2022analyses}.
Furthermore, PINNs have been effectively applied in molecular dynamics \cite{islam2021extraction}, electromagnetics \cite{kovacs2022conditional, fang2021high}, geosciences \cite{haghighat2021sciann, smith2022hyposvi}, and in designing thermal systems \cite{hennigh2020nvidia, cai2021physics}.

Although recent studies have showcased some empirical success with PINNs,  they also highlight several training pathologies. These include spectral bias \cite{rahaman2019spectral, wang2021eigenvector}, unbalanced back-propagated gradients \cite{wang2021understanding, wang2022and}, and causality violation \cite{wang2022respecting}, all of which represent open areas for research and methodological development. To address these issues, numerous studies have focused on enhancing PINNs' performance by refining neural network architectures and training algorithms. Notable efforts are loss re-weighting schemes \cite{wang2021understanding, wang2022and, mcclenny2020self, maddu2022inverse} and adaptive resampling of collocation points, such as importance sampling \cite{nabian2021efficient}, evolutionary sampling \cite{daw2022rethinking}, and residual-based adaptive sampling \cite{wu2023comprehensive}. Simultaneously, significant strides have been made in developing new neural network architectures to improve PINNs' representational capacity, including adaptive activation functions \cite{jagtap2020adaptive}, positional embeddings \cite{liu2020multi, wang2021eigenvector}, and innovative architectures \cite{wang2021understanding, sitzmann2020implicit, gao2021phygeonet, fathony2021multiplicative,moseley2021finite, kang2022pixel}. Further explorations include alternative objective functions, such as those employing numerical differentiation techniques \cite{chiu2022can, huang2024efficient} and variational formulations inspired by Finite Element Methods (FEM) \cite{kharazmi2021hp,patel2022thermodynamically}, along with additional regularization terms to accelerate PINNs convergence \cite{yu2022gradient,son2021sobolev}. The evolution of training strategies also forms an active area of research, with sequential training \cite{wight2020solving, krishnapriyan2021characterizing} and transfer learning \cite{desai2021one, goswami2020transfer,chakraborty2021transfer} showing promise in accelerating learning and improving accuracy.

Despite significant progress in recent years, most existing works on PINNs tend to employ small, and shallow network architectures, leaving the vast potential of deep networks largely untapped. To bridge this gap, we put forth a novel class of architectures coined Physics-Informed residual adaptive networks (PirateNets). Our main contributions are summarized as follows:
\begin{itemize}[leftmargin=*]
    \item We argue that the capacity of PINNs to minimize  PDE residuals is determined by the capacity of network derivatives.
    
    \item We support this argument by proving that, for second-order linear elliptic and parabolic PDEs, the convergence in training error leads to the convergence in the solution and its derivatives.
    
    \item We empirically and theoretically reveal that conventional initialization schemes (e.g., Glorot or He initialization) result in problematic initialization of MLP derivatives and thus worse trainability.
    
    \item We introduce PirateNets to address the issue of pathological initialization, enabling stable and efficient scaling of PINNs to utilize deep networks. The proposed physics-informed initialization in PirateNets also serves as a novel method for integrating physical priors at the model initialization stage.
    
    \item We conduct comprehensive numerical experiments demonstrating that PirateNets achieve consistent improvements in accuracy, robustness, and scalability across various benchmarks.
\end{itemize}

The rest of this paper is organized as follows. 
We first give a concise overview of physics-informed neural networks in Section \ref{sec: pinns}, following the original formulation of Raissi \emph{et al.} \cite{raissi2019physics}.  In Section \ref{sec:init_pathology}, we take the Allen-Cahn equation as a motivating example to illustrate the growing instability in training PINNs with deep neural networks and then propose a potential rationale (mechanism):
\emph{if a PINN model is trained sufficiently, then both its predictions and derivatives should well approximate the solution of the underlying PDE and its derivatives, respectively}. We theoretically justify this claim for linear elliptic and parabolic PDEs and, by investigating the capabilities of MLP derivatives, find that deep MLP derivatives often suffer from ill-posed initializations, which could result in an unstable training process. To address the aforementioned issues, in Section \ref{sec: pirate} we propose a novel framework of PirateNets.
In Section \ref{sec: results}, comprehensive numerical experiments are performed to validate the proposed architecture. In Section \ref{sec: discussion}, we conclude this work with a discussion of potential future research directions.




\section{Physics-informed neural network}
\label{sec: pinns}

In this section, we will briefly overview the standard formulation of PINNs, following the one in \cite{raissi2019physics}.  
Without loss of generality, we consider an abstract PDE of the parabolic form:
\begin{align}
\label{eq: PDE}
     \mathbf{u}_t +  \mathcal{D}[\mathbf{u}] = \mathbf{f}\,,
\end{align} 
defined on a spatial-temporal domain $[0, T] \times \Omega \subset \R^{1 + d}$, where $\Omega$ is a bounded domain in $\R^d$ with regular enough boundary $\p \Omega$, $\mathcal{D}[\cdot]$ is a linear or nonlinear differential operator, and $\mathbf{u}(t,\mathbf{x})$ denotes a unknown solution. 
%
The general initial and boundary conditions can be then formulated as: 
\begin{align}
     \label{eq: IC}
     &\mathbf{u}( 0, \mathbf{x})=\mathbf{g}(\mathbf{x}), \quad \mathbf{x} \in \Omega\,, \\
      \label{eq: BC}
     &\mathcal{B}[\mathbf{u}] = 0, \quad   t\in [0, T], \  \mathbf{x} \in  \partial \Omega\,.
\end{align}
Here, $\mathbf{f}$ and $\mathbf{g}(\mathbf{x})$ are given functions with certain regularity; $\mathcal{B}[\cdot]$ denotes an abstract boundary operator, representing various boundary conditions such as Dirichlet, Neumann, Robin, and periodic conditions.

We aim at approximating the unknown solution $\mathbf{u}(t, \mathbf{x})$ by a deep neural network $\mathbf{u}_{\mathbf{\theta}}(t, \mathbf{x})$, where $\mathbf{\theta}$ denotes the set of all trainable parameters of the network (e.g., weights and biases). If a smooth activation function is employed, $\mathbf{u}_{\mathbf{\theta}}$ provides a  smooth representation that can be queried for any $(t, \mathbf{x})$. Moreover, all required gradients with respect to input variables or network parameters $\mathbf{\theta}$ can be computed via automatic differentiation \cite{griewank2008evaluating}. This allows us to define the interior PDE residuals as
\begin{align}
    \label{eq: pde_residual}
    \mathcal{R}_{{\rm int}}[\mathbf{u}_{\mathbf{\theta}}](t, \mathbf{x}) = \frac{\partial \mathbf{u}_{\mathbf{\theta}}}{\partial t}(t, \mathbf{x}) + \mathcal{D}[\mathbf{u}_{\mathbf{\theta}}](t, \mathbf{x})- \mathbf{f}(\bb{x})\,, \quad (t,\bb{x}) \in [0,T] \times \Omega\,, 
\end{align}
and spatial and temporal boundary residuals, respectively, by 
\begin{align} \label{eq:bcc}
     \mathcal{R}_{{\rm bc}}[\mathbf{u}_{\mathbf{\theta}}](t, \mathbf{x}) = \mathcal{B}[\mathbf{u}_\theta](t, \mathbf{x})\,, \quad (t,\bb{x}) \in [0,T] \times \p \Omega\,, 
\end{align}
and 
\begin{align} \label{eq:icc}
     \mathcal{R}_{{\rm ic}}[\mathbf{u}_{\mathbf{\theta}}]( \mathbf{x}) = \bb{u}_\theta(0,\bb{x}) - \bb{g}(\bb{x})\,, \quad \bb{x} \in \Omega\,.
\end{align}
Then, we train a physics-informed model by minimizing the following composite \emph{empirical loss}:
\begin{align}
    \label{eq: pinn_loss}
    \mathcal{L}(\mathbf{\theta}) := \underbrace{\frac{1}{N_{ic}} \sum_{i=1}^{N_{ic}} \left|  \mathcal{R}_{{\rm ic}}[\mathbf{u}_{\mathbf{\theta}}](\mathbf{x}_{ic}^i) \right|^2}_{\mathcal{L}_{ic}(\mathbf{\theta})} + \underbrace{\frac{1}{N_{bc}} \sum_{i=1}^{N_{bc}} \left| \mathcal{R}_{{\rm bc}}[\mathbf{u}_{\mathbf{\theta}}]( t_{bc}^i, \mathbf{x}_{bc}^i) \right|^2}_{ \mathcal{L}_{bc}(\mathbf{\theta})} + \underbrace{\frac{1}{N_r} \sum_{i=1}^{N_r} \left|  \mathcal{R}_{\rm int}[\mathbf{u}_{\mathbf{\theta}}](t^i_r, \mathbf{x}^i_r) \right|^2}_{\mathcal{L}_r(\mathbf{\theta})}\,,
\end{align}
which aims to enforce the neural network function $\bb{u}_\theta$ to satisfy the PDEs \eqref{eq: PDE} with initial and spatial boundary conditions \eqref{eq: IC}--\eqref{eq: BC}. 
The training data points $\{\mathbf{x}_{ic}^i\}_{i=1}^{N_{ic}}$, $\{t_{bc}^i, \mathbf{x}_{bc}^i\}_{i=1}^{N_{bc}}$ and $\{t_{r}^i, \mathbf{x}_{r}^i\}_{i=1}^{N_{r}}$ can be the vertices of a fixed mesh or points randomly sampled at each iteration of a gradient descent algorithm. We finally remark that although we only review the case of parabolic-type equations here, similar discussions can also be applied to general elliptic and hyperbolic (linear or nonlinear) equations. 

\section{Initialization pathologies in PINNs}
\label{sec:init_pathology}

It has been known for some time that training PINNs is notably more difficult than the supervised training of conventional neural networks. As demonstrated by Wang {\em et al.}, PINNs face several training pathologies including spectral bias \cite{rahaman2019spectral, wang2021eigenvector}, unbalanced losses \cite{wang2021understanding},  and causality violation \cite{wang2022respecting}. Efforts to mitigate these issues include the use of Fourier feature networks \cite{tancik2020fourier}, self-adaptive weighting schemes \cite{wang2021understanding, wang2022and} and the causal and curriculum training algorithms \cite{wang2022respecting, wight2020solving, krishnapriyan2021characterizing}. However, despite these advances, the practical deployment of PINNs has been largely limited to employing small and shallow networks, typically with 5 layers or less. This is in sharp contrast to the exceptional capabilities showcased by deeper networks in the broader field of deep learning \cite{devlin2018bert, radford2021learning, rombach2022high}.

In the following, we will both empirically and theoretically demonstrate that the trainability of PINNs degrades as the network depth increases, particularly when employing multi-layer perceptrons (MLPs) as the backbone architecture. To this end, let us first focus on the Allen-Cahn equation, a challenging benchmark for conventional PINN models that has been extensively studied in recent literature \cite{wight2020solving,wang2022respecting,es2023optimal,daw2022rethinking,anagnostopoulos2023residual}. For simplicity, we consider the one-dimensional case with a periodic boundary condition: 
\begin{align*}
    &u_{t}-0.0001 u_{x x}+5 u^{3}-5 u=0\,, \quad t \in[0,1]\,,\  x \in[-1,1]\,, \\
    &u(0, x)=x^{2} \cos (\pi x)\,, \\
    &u(t, -1)=u(t, 1)\,, \quad u_{x}(t, -1)=u_{x}(t, 1)\,.
\end{align*}
In our experimental setup, we adhere closely to the training pipeline described in Wang {\it et al.} \cite{wang2023expert}.
Specifically, we employ random Fourier feature networks with 256 neurons in each hidden layer and hyperbolic tangent (Tanh) activation functions. The models are trained with a batch size of $1,024$ over $10^5$ steps of gradient descent using the Adam optimizer \cite{kingma2014adam}. We set the initial learning rate to $10^{-3}$, and use an exponential decay rate of $0.9$  every $2,000$ steps. 
In particular, the training of our models incorporates learning rate annealing \cite{wang2021understanding,wang2022respecting} and causal training algorithms \cite{wang2022respecting, wang2023expert} to enhance the model performance and robustness.

\begin{figure}
    \centering
    \includegraphics[width=0.5\textwidth]{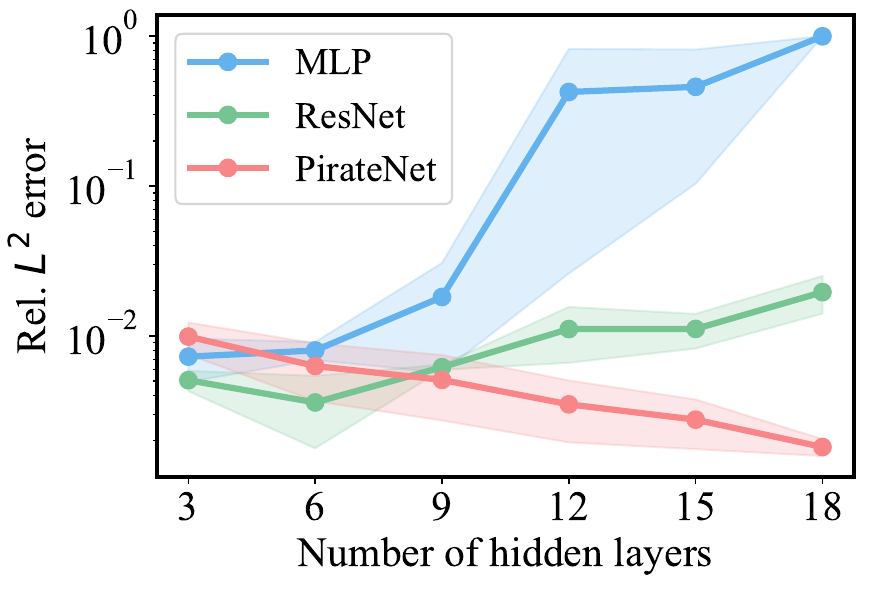}
    \caption{{\em Allen-Cahn equation:} Relative $L^2$ error of training PINNs using MLP, ResNet, and PirateNet backbones of varying depths, averaged over 5 random seeds for each architecture.}
    \label{fig:AC_sweep}
\end{figure}

Figure \ref{fig:AC_sweep} shows the resulting relative $L^2$ errors when training PINNs with MLPs of varying depths. The result is averaged over 5 different random seeds. It can be observed that the prediction error becomes larger as the network depth increases. Eventually, this error almost reaches $100$\% for PINN models with 18 hidden layers, indicating a complete failure of the model to yield a reasonable solution. 

One may argue that the challenges in training deep neural networks can be effectively resolved by incorporating skip connections \cite{he2016deep}. To test this hypothesis, our study compares the performance of MLP networks with residual connections (ResNets) under identical hyper-parameter settings. As depicted in the same figure, the overall trend remains consistent: the error proportionally increases with network depth, albeit at a slightly slower rate.  This implies that, contrary to expectations, the conventional skip connection fails to address the fundamental issues in training PINNs using deeper networks.  These findings highlight a critical limitation in PINN performance when the network depth exceeds a certain threshold. To address this challenge, we propose a novel network architecture in Section \ref{sec: pirate}, referred to as Physics-Informed Residual AdapTivE Networks (PirateNets), designed for efficient and stable training of deep PINN models.  As can be seen in
Figure \ref{fig:AC_sweep}, PirateNets consistently reduces the predictive error as the network depth increases.

To gain deeper insights into the reasons behind the poor performance of MLPs and ResNets, we next examine the initialization process of PINNs. Due to the variety and nonlinearity in PDE expressions, a direct analysis of the initialization of the PDE residual network (i.e., $\mathcal{R}_{\rm int}[\mathbf{u}_{\mathbf{\theta}}]$ in \eqref{eq: pde_residual}) presents significant challenges. Here, in view of the role of the PDE residual loss in  \eqref{eq: pinn_loss} (i.e., enforce the network to satisfy the underlying PDEs), it is reasonable to expect that if the training loss is small, then the derivatives of the network should also closely approximate those of the actual solution. This motivates us to analyze the derivatives of the network to understand the initialization process, instead of examining the  full residual network graph $\mathcal{R}_{\rm int}[\mathbf{u}_{\mathbf{\theta}}]$. 
To be specific, we put forth the following claim. 

\begin{claim} \label{asp1}
    Let $\mathbf{u}(t, \mathbf{x})$ be the solution to a given PDE of $n$-th order in space and $m$-th order in time, and $\mathbf{u}_{\mathbf{\theta}}(t, \mathbf{x})$ denote a approximate solution by a PINN model with parameters $\mathbf{\theta}$, where $\mathbf{x} = (x_1, x_2, \dots, x_d)$. 
    We refer to a $k$-th order spatial or temporal derivative of $\mathbf{u}_{\mathbf{\theta}}$ as the \emph{MLP derivative network}. Then, if the training loss $\mathcal{L}(\theta)$ converges to $0$, then the MLP derivative networks
    will convergence to corresponding solution derivatives in some sense: 
\begin{align*}
    \frac{\partial^k \mathbf{u}_{\theta}}{\partial x_i^k}(t, \mathbf{x}) \longrightarrow \frac{\partial^k u}{\partial x^k_i}(t, \mathbf{x})\,,\quad \frac{\partial^l \mathbf{u}_{\theta}}{\partial t^l}(t, \mathbf{x}) \longrightarrow \frac{\partial^l u}{\partial t^l}(t, \mathbf{x})\,,
\end{align*}
for some $k = 1, 2, \dots, n$, $l = 1, 2, \dots, m$, and $i = 1, 2, \dots, d$.
\end{claim}

In fact, we show in Propositions \ref{prop: elliptic} and \ref{prop: parabolic} below that the above claim indeed holds for linear elliptic and parabolic PDEs. We remark that the error estimates for PINNs in Sobolev norms have also been explored very recently for various PDE models, such as elliptic, elasticity, parabolic, hyperbolic, and Stokes equations, in a unified manner \cite{zeinhofer2023unified}.

For ease of expositions, we first introduce some notations. Let $\Omega \subset \mathbb{R}^d$ be a bounded open set, and define the partial differential operator: 
\begin{equation} \label{eq:elliopp}
     \mc{D}[u] := - \sum_{i,j= 1}^n (a^{ij}(\mathbf{x}) u_{x_i})_{x_j} + \sum_{i = 1}^n b^i(\mathbf{x})u_{x_i} + c(\mathbf{x}) u\,,
\end{equation}
with smooth coefficients $a^{ij}(\mathbf{x})$, $b^i(\mathbf{x})$, and $c(\mathbf{x})$. We assume that $\mc{D}$ is uniformly elliptic, i.e., for some constant $c > 0$, $\sum_{i,j = 1}^n a^{ij}(\mathbf{x}) \xi_i \xi_j\ge c |\xi|^2$ for a.e. $x \in \Omega$ and all $\xi \in \R^n$. 

We also define $B_r(\mathbf{x})$  by the open ball in $\R^d$ of radius $r$ centered at $\mathbf{x}$ and $Q_r(t,\mathbf{x}) := (t, t + r^2) \times B_r(\mathbf{x})$. For a time--space domain $U \subset \R^{d + 1}$, we denote the set $C^{r, k + 2r}(\bar{U})$ of bounded continuous functions $u(t,x)$ such that their derivatives $\p_t^\rho \p_{\mathbf{x}}^\alpha u$ for $|\alpha| + 2 \rho \le k + 2r$ and $\rho \le r$ are bounded and continuous in $U$, and can be extended to $\bar{U}$. For $u \in C^{r, k + 2r}(\bar{U})$, we define 
\begin{align*}
    \|u\|_{W_2^{r, k+2 r}(U)} := \sum_{\substack{|\alpha|+2 \rho \leq k+2 r \\ \rho \leq r}}\left\|\partial_t^\rho \p_{\bb{x}}^\alpha u\right\|_{L^2(U)}\,.
\end{align*}
Then, the Sobolev space $W^{r,k+2r}_2(U)$ is given by the completion of $C^{r, k + 2r}(\bar{U})$ with respect to $\norm{\cdot}_{W^{r,k+2r}_2(U)}$.

\begin{proposition}
\label{prop: elliptic}
 Consider the second-order elliptic Dirichlet problem:
\begin{equation*}
    \mc{D}[u] = f\,,
\end{equation*}    
with Dirichlet boundary condition and $f \in L^2(\Omega)$. Let $u: \Omega \rightarrow \mathbb{R}$ be its solution and $u_{\mathbf{\theta}}$ be a smooth approximation by PINNs. 
Define the \emph{expected loss} function: 
\begin{align} \label{eq:expected}
     \h{\mc{L}}(\mathbf{\theta}) =  \int_{\partial \Omega} \left|u_\theta \right|^2  \,  \rd \mathbf{x} +  \int_\Omega \left|\mc{D}[u_\theta] - f\right|^2  \, \rd \mathbf{x}\,.
\end{align}
Then, for any compact $V  \subset  \subset \Omega$, there exist a constant $C$ such that
\begin{align} \label{est:ellip}
    \|u - u_{\mathbf{\theta}}\|_{H^2(V)}^2 \leq C  \h{\mc{L}}(\mathbf{\theta})\,.
\end{align}
\end{proposition}

We emphasize that the above proposition is based on the expected loss \eqref{eq:expected}, while the practical training process employs the empirical loss as in \eqref{eq: pinn_loss}:
\begin{equation*}
     \mathcal{L}(\mathbf{\theta}) = \frac{1}{N_{bc}} \sum_{i=1}^{N_{bc}} \left| u_{\mathbf{\theta}}( \mathbf{x}_{bc}^i) \right|^2 + \frac{1}{N_r} \sum_{i=1}^{N_r} \left| \mc{D}[u_\theta](t^i_r, \mathbf{x}^i_r) - f(t^i_r, \mathbf{x}^i_r) \right|^2\,.
\end{equation*}
In general, we have that when the numbers of randomly sampled training points $N_{bc}$ and $N_r$ tend to infinity, $\mathcal{L}(\mathbf{\theta})$ converges to 
the corresponding $\h{\mc{L}}(\mathbf{\theta})$. We refer the interested readers to \cite[Section 2.4.1]{mishra2022estimates} for the detailed analysis. This means that if the network is sufficiently trained, i.e., the loss $\mc{L}(\theta)$ is small, the PINN solution $u_\theta$ can locally approximate $u$ in the $H^2$ norm, which, by Sobolev embedding theorem, further implies that in the case of $d = 1$, the derivative $\tfrac{d}{dx} u_\theta(x)$ approximates $\tfrac{d}{dx} u(x)$ pointwisely. A similar result can be found in \cite[Lemma 4.1]{zeinhofer2023unified}. We next state the analogous result for parabolic PDEs, which complements the discussions in \cite[Section 4.5]{zeinhofer2023unified}. 




\begin{proposition}
\label{prop: parabolic} 
Consider a second-order parabolic equation with Dirichlet boundary condition:
\begin{align*}
    & u_t + \mc{D}[u] = f\,,\quad [0,T] \times \Omega\,,   \\
    & u = 0\,, \quad   [0,T] \times \p \Omega\,,  \\
    & u = g\,, \quad  \{t = 0\} \times  \Omega\,,
\end{align*}
with $f \in L^2([0,T]; L^2(\Omega))$ and $g \in L^2(\Omega)$. 
Let $u: [0,T] \times \Omega \rightarrow \mathbb{R}$ be its solution and $u_{\mathbf{\theta}}$ be a smooth approximation by PINNs. 
Define the \emph{expected loss}: 
\begin{align*}
    \h{\mc{L}}(\mathbf{\theta}) = \int_\Omega |\mc{R}_{\rm int}[u_\theta](\bb{x})|^2 \, \rd x  + \int_{0}^T \int_{\partial \Omega} |u_\theta(t,\mathbf{x})|^2  \, \rd t \rd \mathbf{x} + \int_{0}^T \int_\Omega |\mc{R}_{\rm int}[u_\theta](t,\mathbf{x})|^2   \, \rd t \rd \mathbf{x}\,,
\end{align*}
with $\mc{R}_{\rm int}[u_\theta]$ and $ \mathcal{R}_{{\rm bc}}[u_\theta]$ given in \eqref{eq: pde_residual} and \eqref{eq:icc}, respectively. Then, for any set $Q_r   (t,\bb{x}) \subset \subset [0,T] \times \Omega$, there exist a constant $C$ such that
\begin{align*}
    \|u - u_{\mathbf{\theta}}\|_{W_2^{1, 2}\left(Q_r\right)}^2  \leq C  \h{\mc{L}}(\mathbf{\theta})\,.
\end{align*}
\end{proposition}

The proofs of Propositions \ref{prop: elliptic} and \ref{prop: parabolic}  can be found in Appendix \ref{app:a}, which are essentially the applications of interior estimates of PDEs. Recalling from \cite{hormander1958interior} that the solutions of a formally hypoelliptic equation $P(x, D)[u] = f$ are infinitely differentiable if the coefficients in the differential operator $P(x,D)$ and $f$ are infinitely differentiable, one may expect similar results hold for a fairly large class of equations. However, for some evolution equations, e.g., wave equation and Burgers equation, the singularity of the initial data can propagate into the interior of the domain so that the $L^2$ residual loss would be insufficient to control the higher-order derivative of the solution, and hence some Sobolev loss may need to be used \cite{yu2022gradient, son2021sobolev}. The following proposition discusses the case of the second-order hyperbolic equation, from which we see that in order to obtain the convergence of the first-order derivative of the solution, the $H^1$ temporal boundary residual is employed. Its proof follows from standard regularity result of hyperbolic equations \cite[Section 7.2, Theorem 5]{evans2022partial}; see also \cite[Theorem 4.1]{lasiecka1986non}. 

\begin{corollary}
Consider the hyperbolic initial/boundary-value problem: 
\begin{align*}
    & u_{tt} + \mc{D}[u] = f\,,\quad [0,T] \times \Omega\,,   \\
    & u = 0\,, \quad   [0,T] \times \p \Omega\,,  \\
    & u = g\,,\ u_t = h\,, \quad  \{t = 0\} \times  \Omega\,,
\end{align*}
with $f \in L^2([0,T]; L^2(\Omega))$, $g \in H_0^1(\Omega)$, and $h \in L^2(\Omega)$.  Let $u$ be its solution and $u_{\mathbf{\theta}}$ be a smooth approximation by PINNs. Define the \emph{expected loss}: 
\begin{align*}
    \h{\mc{L}}(\mathbf{\theta}) = &  \norm{(u_{\theta})_{tt} + \mc{D}[u_\theta] - f}^2_{L^2([0,T]; L^2(\Omega))} +  \norm{(u_{\theta})_t - h}^2_{L^2(\Omega)} \\ & + \norm{u_\theta(0,\cdot) - g}^2_{H^1(\Omega)}   + \norm{u_\theta}^2_{L^2([0,T]; H^1(\p \Omega))} + \norm{u_\theta}^2_{H^1([0,T]; L^2(\p \Omega))}\,.
\end{align*}
Then, we have for some constant $C > 0$, 
 \begin{align*}
     \sup_{0 \le t \le T} (\norm{u_\theta - u}^2_{H^1(\Omega)} + \norm{u_t}^2_{L^2(\Omega)}) \le C   \h{\mc{L}}(\mathbf{\theta})\,.
 \end{align*}
\end{corollary}

With the above discussions, we narrow our focus on the analysis of the network derivatives. Without loss of generality, we consider an MLP with scalar inputs and outputs. Specifically, let $x \in \R $ be the input coordinate, $\mathbf{g}^{(0)}(x) = x$ and $d_0 = d_{L+1} = 1$. An MLP $u_{\mathbf{\theta}}(x)$ is recursively defined as follows
\begin{align}
    \label{eq: mlp_1}
    \mathbf{u}^{(l)}_{\mathbf{\theta}}(x) = \mathbf{W}^{(l)} \cdot \mathbf{g}^{(l-1)}(x) + \mathbf{b}^{(l)}\,, \quad \mathbf{g}^{(l)}(x) = \sigma(\mathbf{u}_\theta^{(l)}(x))\,, \quad l = 1,2, \dots, L\,,
\end{align}
with a final linear layer defined by
\begin{align}
    \label{eq: mlp_2}
    u_{\mathbf{\theta}}(x) &= \mathbf{W}^{(L+1)} \cdot \mathbf{g}^{(L)}(x) + \mathbf{b}^{(L+1)}\,,
\end{align}
where $\mathbf{W}^{(l)} \in \R^{d_l \times d_{l-1}}$ is the weight matrix in $l$-th layer and $\sigma$ is an element-wise activation function. Here, $\mathbf{\theta}=\left(\mathbf{W}^{(1)}, \mathbf{b}^{(1)}, \ldots, \mathbf{W}^{(L+1)},  \mathbf{b}^{(L+1)}\right)$ 
represents all trainable parameters of the network. In particular, we suppose that the network is equipped with Tanh activation functions and all weight matrices are initialized by the Glorot initialization scheme \cite{glorot2010understanding},  i.e., $\mathbf{W}^{(l)} \sim \mathcal{D}(0, \frac{2}{d_{l} + d_{l-1}})$ for $l=1,2,\dots, L+1$. Moreover, all bias parameters are initialized to zeros. These are standard practices in the vast majority of existing PINN implementations. In this setting, we have the following proposition with the proof given in Appendix \ref{app:a}. 

\begin{proposition} 
\label{prop}

Given the above setup, the following holds at initialization of an MLP $u_{\mathbf{\theta}}(x)$.
\begin{enumerate}[label=(\alph*)]   
    \item The first-order derivative of $u_{\mathbf{\theta}}$  is given by
    \small
\begin{align}
    \frac{\partial u_{\mathbf{\theta}}}{\partial x}(x) 
    & = \prod_{i=L+1}^2\left(\mathbf{W}^{(i)} \cdot \operatorname{diag}\left(\dot{\sigma}\big(\mathbf{u}_\theta^{(i-1)}(x)\big)\right) \right) \cdot \mathbf{W}^{(1)}\,.
    \label{eq: mlp_deriv}
\end{align}
   Furthermore, assuming that the MLP lies in a linear regime, i.e., $\sigma(x) \approx x $ and $\dot{\sigma}(\cdot) \approx 1$,  it follows that
    \begin{align}
         \frac{\partial u_{\mathbf{\theta}}}{\partial x}(x) \approx \mathbf{W}^{(L+1)} \cdot \mathbf{W}^{(L)}  \cdots \cdot \mathbf{W}^{(1)}\,.
    \end{align}

    \item If all hidden layers of the MLP have the same number of neurons, i.e., $d_1 = d_2 = \cdots = d_L = d$, then 
    \begin{align}
        \operatorname{Var}\left( \frac{\partial u_{\mathbf{\theta}}}{\partial x}(x)\right) \lesssim \frac{1}{d}\,.
    \end{align}
    Consequently, for any $\epsilon > 0$,
    \begin{align}
        \mathbb{P} \left( \left| \frac{\partial u_{\mathbf{\theta}}}{\partial x}(x) \right| > \epsilon \right) \lesssim \frac{1}{d \epsilon^2}\,.
    \end{align}
\end{enumerate}

\end{proposition}

From this proposition, it becomes evident that, in contrast to the conventional forward pass of MLPs where the weight matrices are hidden inside the activation functions,  all weight matrices directly contribute to the final output of $\frac{\partial u_{\mathbf{\theta}}}{\partial x}$.  Also note that, under a linear regime$,  \frac{\partial u_{\mathbf{\theta}}}{\partial x}$ behaves as a deep linear network at initialization.  Such networks have limited expressivity and are more vulnerable to vanishing and exploding gradients. More critically, this expression is independent of the input coordinates at initialization, therefore preventing $ \frac{\partial u_{\mathbf{\theta}}}{\partial x}$ from effectively approximating the solution derivatives in the early stage of training.

To gain further insight, note that the variance of $\frac{\partial u_{\mathbf{\theta}}}{\partial x}$ is bounded by the network's width, which is independent of the network depth.  As a consequence, $\frac{\partial u_{\mathbf{\theta}}}{\partial x}$ tends to be closer to a constant in wider MLPs. This implies that the network is closer to a trivial zero solution at the initialization, which is particularly unfavorable for solving homogeneous PDEs. Although our theoretical analysis mainly focuses on first-order derivatives and the Tanh activation function, subsequent numerical evidence suggests that these insights extend to higher-order derivatives and other common activation functions.

\begin{figure}
    \centering
    \includegraphics[width=0.9\textwidth]{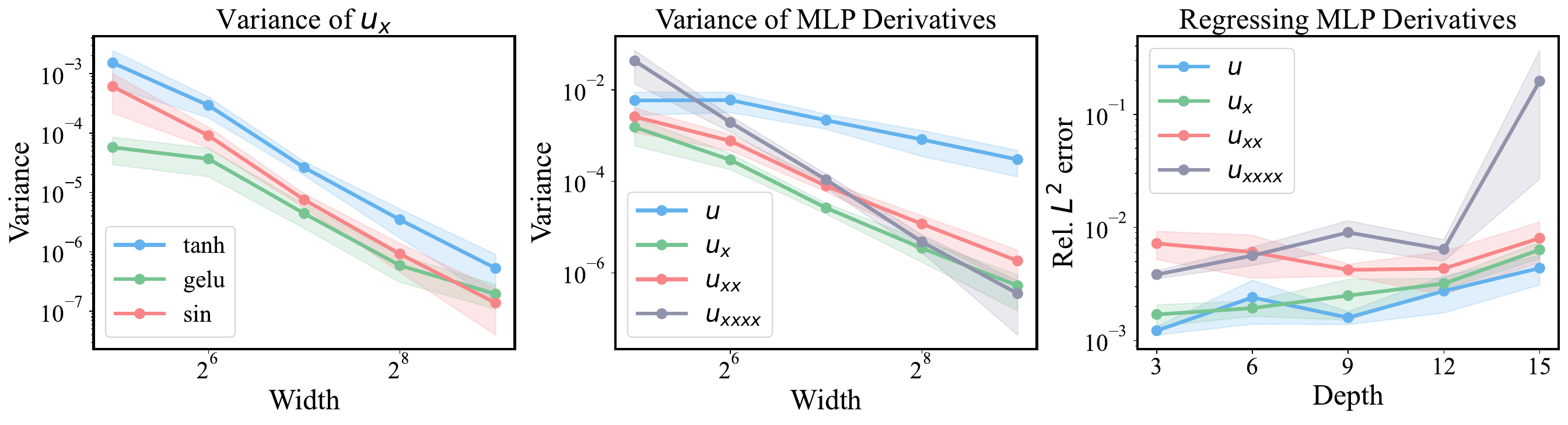}
    \caption{{\em Regression:} {\em Left:} Variance of the network derivative
    equipped with different activations 
    across various network widths at initialization.   {\em Middle:} Variance of MLP derivatives of different orders across various network width at initialization. {\em Right:} Relative $L^2$ error in approximating $y(x) = \sin(2 \pi x)$ with MLP derivatives of different orders. All statistics are averaged over 5 random seeds.}
    \label{fig:regression}
\end{figure}

To illustrate these findings through a concrete numerical test, we consider a simple regression problem for approximating a sinusoidal function $y(x) = \sin(2\pi x ) $ for $x \in [0, 1]$. We generate a training dataset $\{(x_i, y_i)\}_{i=1}^N$ by evaluating the target function $y$ over a uniform grid of size $N=256$. Our goal is to investigate the behavior of MLPs and their associated derivative networks at initialization and during training. First, we fix the depth of the MLP at 3 hidden layers and examine ${\rm Var}\,(\frac{u_{\mathbf{\theta}}}{\partial x})$ for various widths and activations at initialization. As shown on the left panel of Figure  \ref{fig:regression}, the variance decays exponentially with increasing width,  not only for Tanh but also for GeLU and sine activations. Moreover, in the middle panel of Figure \ref{fig:regression}, we show the variance of different order derivatives of the MLP at initialization. It can be seen that the second and fourth-order derivatives of the MLP exhibit a similar trend as those of the first-order derivatives. These observations are consistent with Proposition \ref{prop}. 

Next, we study the capability of MLP derivative networks of different order to approximate the target function. The computational graph of the $k$-th derivative network $\frac{\partial_x^k u_{\mathbf{\theta}} }{\partial x^k}$ can be readily obtained via automatic differentiation, allowing to easily train such networks by minimizing is a standard mean squared error loss given by
\begin{align}
    \mathcal{L}(\mathbf{\theta}) = \frac{1}{N} \sum_{i=1}^N \left| \frac{\partial_x^k u_{\mathbf{\theta}} }{\partial x^k}(x_i) - y_i \right|^2, \quad k=1,2,4\,.
\end{align}
For our experiments,  we fix the base MLP width at 128, employ Tanh activation functions, and study the effect of various network depths on predictive accuracy. All models are trained for $10^4$ steps of full-batch gradient descent using the Adam optimizer.  
The initial learning rate is $10^{-3}$, following an exponential decay with a decay rate of 0.9 every $1,000$ steps. The resulting relative $L^2$ errors are summarized on the right panel of Figure \ref{fig:regression}.  We can see that, in comparison to the MLP output $u$ itself, higher-order derivatives of deeper MLPs tend to yield larger errors. Consequently, we may conclude that the trainablity of MLP derivative networks is worse than the base MLP representing $u_{\mathbf{\theta}}$ itself. This result serves as a proxy to indicate that evaluating and minimizing the outputs of a PDE residual network $\mathcal{R}_{\rm int}[\mathbf{u}_{\mathbf{\theta}}]$ is considerably more difficult and unstable than minimizing a conventional supervised loss for fitting $\mathbf{u}_{\mathbf{\theta}}$ to data. This also emphasizes a critical challenge in designing and optimizing deep and wide PINNs, particularly in terms of their effectiveness in approximating solution derivatives.

\section{Physics-informed residual adaptive networks (PirateNets)}
\label{sec: pirate}

In this section, we introduce Physics-Informed Residual AdapTivE Networks (PirateNets), a novel architecture designed to address the initialization issue elaborated in Section \ref{sec:init_pathology}. Figure \ref{fig:pirate} below illustrates the main modules of the forward pass of a PirateNet. In detail, the input coordinates $\mathbf{x}$ are first mapped into a high dimensional feature space by an embedding function $\Phi(\mathbf{x})$. In this work, we employ random Fourier features \cite{tancik2020fourier}:
\begin{align*}
    \Phi(\mathbf{x})= \begin{bmatrix}
    \cos (\mathbf{B x} ) \\
    \sin (\mathbf{B x} )
    \end{bmatrix},
\end{align*}
where each entry in $\mathbf{B} \in \R^{m \times d}$ is i.i.d. sampled from a Gaussian $\mathcal{N}(0, s^2)$ with the standard deviation $s > 0$ being a user-specified hyper-parameter. Such an embedding has been extensively validated for reducing spectral bias in the training of PINNs, thus enabling a more efficient approximation of high-frequency solutions \cite{wang2021eigenvector}.  

Then, the embedded coordinates $\Phi(\mathbf{x})$ are sent into two dense layers: 
\begin{align*}
    \mathbf{U} &= \sigma(\mathbf{W}_1 \Phi(\mathbf{x}) + \mathbf{b}_1  )\,, \quad
    \mathbf{V} = \sigma(\mathbf{W}_2 \Phi(\mathbf{x}) + \mathbf{b}_2  )\,,
\end{align*}
where $\sigma$ denotes a point-wise activation function. These two encoding maps act as gates in each residual block of the architecture. 
This step is motivated by \cite{wang2021understanding} and has been widely used to enhance the trainability of MLPs and improve the convergence of PINNs \cite{wang2023expert,anagnostopoulos2023residual}.

Let $\mathbf{x}^{(l)}$ denote the input of the $l$-th block for $1 \le  l \le L$. The forward pass in each PirateNet block is defined by the following iterations:
\begin{align}
    \mathbf{f}^{(l)}  &= \sigma\big(\mathbf{W}^{(l)}_1 \mathbf{x}^{(l)} + \mathbf{b}^{(l)}_1\big)\,, \label{eq: step1}  \\
    \mathbf{z}^{(l)}_1 &= \mathbf{f}^{(l)} \odot \mathbf{U} + (1 - \mathbf{f}^{(l)}) \odot \mathbf{V}\,,  \label{eq: gate1} \\
     \mathbf{g}^{(l)}  &= \sigma\big(\mathbf{W}^{(l)}_2 \mathbf{z}_1^{(l)} + \mathbf{b}^{(l)}_2\big)\,, \\
     \mathbf{z}^{(l)}_2 &= \mathbf{g}^{(l)} \odot \mathbf{U} + (1 - \mathbf{g}^{(l)}) \odot \mathbf{V}\,,  \label{eq: gate2} \\
      \mathbf{h}^{(l)}  &= \sigma\big(\mathbf{W}^{(l)}_3 \mathbf{z}_2^{(l)} + \mathbf{b}^{(l)}_3\big)\,, \\
    \mathbf{x}^{(l+1)} &= \alpha^{(l)}  \mathbf{h}^{(l)} + (1 - \alpha^{(l)})   \mathbf{x}^{(l)}\,,
    \label{eq: skip}
\end{align}
where $\odot$ denotes a point-wise multiplication and $\alpha^{(l)} \in \mathbb{R}$ is a trainable parameter. All the weights are initialized by the Glorot scheme \cite{glorot2010understanding}, while biases are initialized to zero.

The final output of a PirateNet of $L$ residual blocks is given by
\begin{align}
    \mathbf{u}_{\mathbf{\theta}} = \mathbf{W}^{(L+1)} \mathbf{x}^{(L)}\,.
\end{align}

It is worth noting from \eqref{eq: step1}--\eqref{eq: skip} that each residual block consists of three dense layers with two gating operations, followed by an adaptive residual connection across the stacked layers. Thus, a PirateNet of $L$ residual blocks will have a depth of $3L$ and its total number of trainable parameters is comparable to that of an MLP of the same depth.

A key aspect of PirateNets is the trainable parameters  $\alpha^{(l)}$ in the skip connections, which determine the nonlinearity of the $l$-th block. Specifically, when $\alpha^{(l)}=0$, $\mathbf{x}^{(l+1)} = \mathbf{x}^{(l)}$ by \eqref{eq: skip}, meaning that in this case, the $l$-th block is an identity map. In contrast, when $\alpha^{(l)} = 1$, the mapping becomes fully nonlinear without any shortcuts. Throughout all experiments in this work,  we initialize $\alpha^{(l)}$ to zero for all blocks, leading to the final output of a PirateNet $\mathbf{u}_{\mathbf{\theta}}$ being a linear combination of the first layer embeddings at initialization, i.e.,
\begin{align}
\label{eq: linear}
\mathbf{u}_{\mathbf{\theta}}(\mathbf{x}) = \mathbf{W}^{(L+1)}\Phi(\mathbf{x})\,.
\end{align}
By doing so, we circumvent the initialization pathologies of deep networks discussed in Section \ref{sec:init_pathology}. PirateNets are designed to learn the requisite nonlinearity encoded in $\alpha$, based on the PDE system, during the training process. As a result, both their trainability and expressivity are restored, with the models' forward pass becoming more nonlinear and deeper. In fact, the PDE solution can be simply approximated by a small shallow network or even a linear combination of some basis, similar to spectral and finite element methods.  The rationale behind employing deep neural networks lies in leveraging additional nonlinearities to minimize the PDE residuals,  thereby enabling the network to learn the solution and its derivatives accordingly.

From equation \eqref{eq: linear}, another key observation is that PirateNets can be viewed as a linear combination of basis functions at initialization. This not only allows one to control the inductive bias of the network by appropriate choice of basis, but also enables the integration of various types of existing data into the initialization phase of the network. Specifically, given a set of solution measurements, denoted as $\mathbf{Y} = \{y_i \}_{i=1}^n$, then one can readily initialize the last linear layer of the model by the following least square problem:
\begin{align}
    \label{eq: lq}
    \min_{\mathbf{W}} \left\| \mathbf{W} \Phi  - \mathbf{Y} \right\|_2^2\,.
\end{align}
As a result, PirateNets offers an optimal initial guess, based on the available data, in the $L^2$ sense. It is important to note that the data for this initialization could come from a variety of sources, including experimental measurements, initial and boundary conditions, solutions derived from surrogate models, or by approximating the solution of the linearized PDE.
Indeed, we can apply the same initialization procedure to any network architecture with a linear final layer (see Equation \eqref{eq: mlp_2}), while a major consideration is the potentially limited expressivity of randomly initialized basis to accurately fit the data. Taken together, our proposed approach paves a new way to incorporate physical priors into machine learning pipelines through appropriate network initialization.

\paragraph{Connections to Prior Work:} It is important to acknowledge that the adaptive residual connection proposed in our work is not entirely novel. Similar concepts have been previously introduced by Bresson {\em et al.}\cite{savarese2017residual}, who proposed the so-called \emph{Residual Gates} as follows:
\begin{align}
    \mathbf{y} &= \text{ReLU}(\alpha) \cdot (\mathcal{F}(\mathbf{x}) + \mathbf{x}) + (1 - \text{ReLU}(\alpha)) \cdot  \mathbf{x}\,, \notag \\
     &= \text{ReLU}(\alpha) \cdot \mathcal{F}(\mathbf{x}) + \mathbf{x}\,,
     \label{eq: res_gate}
\end{align} 
where $\mathcal{F}$ represents a residual function and $\mathbf{x}$ and $ \mathbf{y}$ denote the input and output of a layer, respectively. While our approach is similar, a subtle yet fundamental difference lies in our initialization $\alpha = 0$ in Equation \eqref{eq: skip}, as opposed to their initialization $\alpha = 1$ in Equation \eqref{eq: res_gate}. Consequently, Bresson {\em et al.}'s model is initialized as an original ResNet, while ours is initialized as an identity mapping, with the purpose of overcoming the initialization pathologies in PINNs. The significant impact of this initialization choice is clearly illustrated in the ablation studies presented in Figures \ref{fig:gs_sweep} and \ref{fig:gl_sweep}, which confirm the necessity of initializing $\alpha$ as zero for achieving optimal performance in PINNs.
Additionally, He {\em et al.} \cite{he2016deep, he2016identity} has extensively explored various residual connection variants. These typically place the activation after the addition. In contrast, PirateNets requires placing the activation before the addition to ensure an exact identity mapping throughout the entire forward pass by setting $\alpha = 0$ at initialization.

Our proposed initialization technique also shares similarities with the Physics-informed Extreme Learning Machine (PIELM) \cite{dwivedi2020physics,dong2021local}.  Both methods focus on initializing the output layer weights using the least squares method. However, PIELM is designed to solve the PDE in a single step without a training process, leading to three important limitations. First, PIELM typically employs shallow randomly initialized networks, potentially limiting their capacity to represent complex solutions effectively. Second, tackling nonlinear problems becomes more challenging due to the nonlinearity of the associated least squares problem, which is notably sensitive to the initial guess and susceptible to local minima. Third, the application of PIELM is limited as it cannot be used for inverse problems. In contrast, PirateNets do not attempt to solve the PDE in one shot.  Instead, our goal is to use a linear least squares approach to initialize the model as the best possible approximation based on our available knowledge and/or data.  We expect PirateNets to correct and learn the solution during the training phase.  An advantage of our method is its simplicity and efficiency in solving the linear least squares problem, which guarantees a global minimum and requires minimal computational resources. Moreover, the flexibility in choosing an appropriate embedding mapping ensures accurate data reconstruction, which greatly benefits from the dedicated design of the adaptive skip connections in PirateNets.

\paragraph{Remark} In PirateNets, random Fourier features are our primary choice for constructing coordinate embeddings. However, it is worth mentioning that there are other feasible options, such as Chebyshev polynomials, radial basis functions, etc. Another alternative is the spectral representation, where frequencies are uniformly sampled on a grid. Despite not being detailed in this paper, we have explored this type of embedding as well. Interestingly, our experiments reveal that compared to random Fourier features, this spectral representation can reconstruct the data with greater accuracy. Yet, it also showed a greater tendency towards overfitting, characterized by a decrease in loss but an increase in the test error. Therefore, we choose to use random Fourier features for the coordinate embedding in PirateNets.

\begin{figure}
    \centering
    \includegraphics[width=1.0\textwidth]{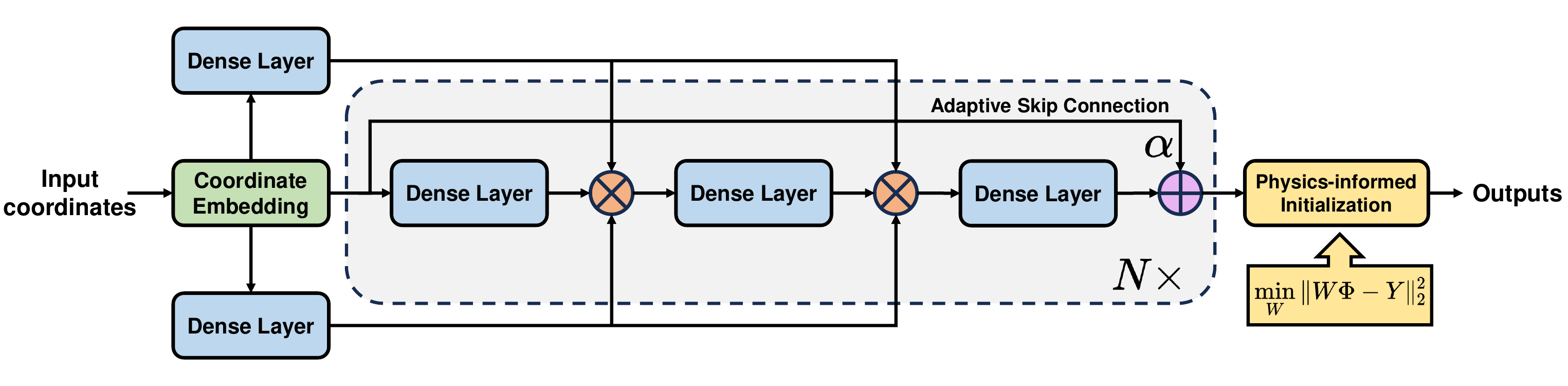}
    \caption{{\em Physics-informed residual adaptive networks (PirateNets):} In our model, input coordinates are first projected into a high-dimensional feature space using random Fourier features, then followed by passing through $N$ adaptive residual blocks. Each block consists of three dense layers, augmented with two gating operations that incorporate shallow latent features. The key module of the architecture is the adaptive skip connection with a trainable parameter $\alpha$ initialized at $0$, so that at the initialization phase, each block reduces to an identity mapping, and the model can be viewed as a linear combination of the coordinate embeddings. It turns out that this approach helps to circumvent the issue of pathological initialization in deep PDE residual networks. We propose a physics-informed initialization for the final layer by solving a least squares problem to fit the available data, while all other weights are initialized following the Glorot scheme, and biases are set to zero. As training progresses, the depth of the model gradually increases as the nonlinearities become activated, enabling the model to progressively recover its approximation capacity.}
    \label{fig:pirate}
\end{figure}


\section{Results}
\label{sec: results}


In this section, we demonstrate the effectiveness of the proposed PirateNet architecture across a diverse collection of benchmarks. To establish a strong baseline,  we adopt the training pipeline and recommended hyper-parameters from Wang \emph{et al.} \cite{wang2023expert}. Specifically, we employ random Fourier features \cite{tancik2020fourier} as coordinate embedding and  Modified MLP \cite{wang2021understanding,wang2023expert} of width 256  as the baseline architecture backbone. 
Additionally, all weight matrices are enhanced using random weight factorization (RWF) \cite{wang2022random}. The Tanh activation function is employed by default, unless specified otherwise. Exact periodic boundary conditions are applied as required \cite{dong2021method}.

For model training, we use mini-batch gradient descent with the Adam optimizer \cite{kingma2014adam}, where collocation points are randomly selected in each iteration. Our learning rate schedule includes an initial linear warm-up phase of $5,000$ steps, starting from zero and gradually increasing to  $10^{-3}$, followed by an exponential decay at a rate of $0.9$. Following the best practices \cite{wang2023expert}, we also employ a learning rate annealing algorithm \cite{wang2021understanding,wang2023expert} to balance losses and causal training \cite{wang2022respecting,wang2023expert} to mitigate causality violation in solving time-dependent PDEs.


For all examples, we compare the proposed PirateNet against our baseline under exactly the same hyper-parameter settings. Our main results are summarized in Table \ref{tab: sota}.  For ease of replication, we detail all hyper-parameters used in our experiments in Tables \ref{tab: ac_config}, \ref{tab: kdv_config}, \ref{tab: gs_config}, \ref{tab: gl_config}, and \ref{tab: ldc_config}.
For the physics-informed initialization, we employ the   \texttt{jax.numpy.linalg.lstsq} routine to solve the associated least square problem, which returns the solution with the smallest 2-norm. 
In cases where the data comes from an initial condition $u_0(x)$, we initialize the weights of the last layer to fit  $u_{\mathbf{\theta}}(t, x)$ to $u_0(x)$ for all $t$. We find that this approach leads to a more stable training process and better results than only fitting the initial conditions at $t=0$.

Moreover, we conduct comprehensive ablation studies to validate the proposed network architecture and its components.  Specifically, these studies focus on: (a) the scalability of PirateNet, (b) the functionality of adaptive residual connection, (c) the efficiency of gating operators, and (d) the effectiveness of physics-informed initialization.  For each ablation study, it is worth emphasizing that all compared models are trained under exactly the same hyper-parameter settings and the results are averaged over five random seeds. 
It is also noteworthy that the computational cost of training PirateNet is comparable to that of training a Modified MLP of equivalent depth. For a more detailed quantitative analysis, readers are referred to Wang {\em et al.} \cite{wang2023expert}. The code and data for this study will be made publicly available at \url{https://github.com/PredictiveIntelligenceLab/jaxpi}.

\begin{table}
    \centering
    \begin{tabular}{lcc} 
        \toprule
            \textbf{Benchmark} & \multicolumn{1}{c}{\textbf{PirateNet }} & \multicolumn{1}{c}{\textbf{JAX-PI} \cite{wang2023expert}} \\ 
        \midrule
        Allen-Cahn  & $\mathbf{2.24 \times 10^{-5}}$ &  \num{5.37e-5} \\ 
        Korteweg–De Vries  &$\mathbf{4.27 \times 10^{-4}}$ & \num{1.96e-3} \\
         Grey-Scott & $\mathbf{3.61 \times 10^{-3}}$ & ${6.13} $ \\
        Ginzburg-Landau  & $\mathbf{1.49 \times 10^{-2}}$ & $3.20 \times 10^{-2}$  \\
        Lid-driven cavity flow (Re=3200) & $\mathbf{4.21 \times 10^{-2}} $ &  ${1.58 \times 10^{-1}} $ \\
        \bottomrule
    \end{tabular}
    \caption{State-of-the-art relative $L^2$ test error for various PDE benchmarks using the PirateNet architecture.}
    \label{tab: sota}
\end{table}


\subsection{Allen-Cahn equation}
\begin{figure}
     \centering
     \begin{subfigure}[b]{0.9\textwidth}
         \centering
         \includegraphics[width=\textwidth]{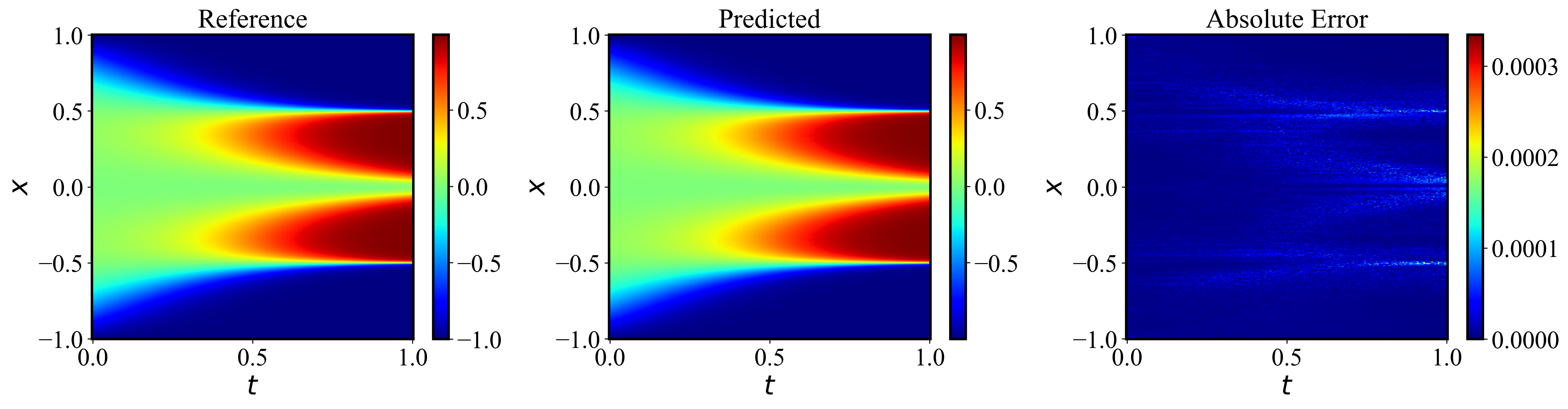}
         \label{fig:ac_pred}
     \end{subfigure}
     \hfill
     \begin{subfigure}[b]{0.9\textwidth}
         \centering
         \includegraphics[width=\textwidth]{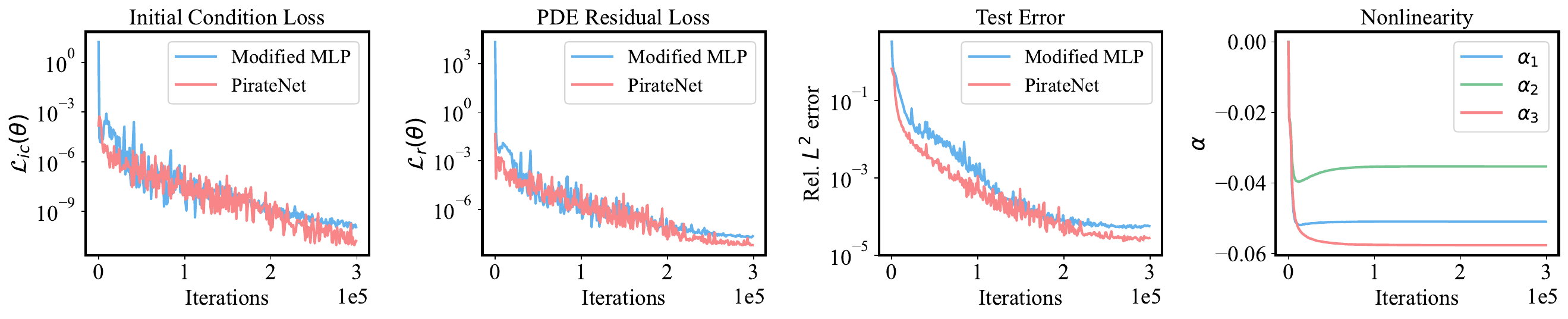}
         \label{fig:ac_pirate_stat}
     \end{subfigure}
        \caption{{\em Allen-Cahn equation:} {\em Top:} Comparison between the solution predicted by a trained PirateNet and the reference solution. The detailed hyper-parameter settings are presented in Table \ref{tab: ac_config}.
        {\em Bottom:}   Convergence of the initial condition loss, the PDE residual loss, and the relative $L^2$ test error during the training of a PirateNet and a Modified MLP backbone, alongside the evolution of nonlinearities in each of the PirateNet residual blocks. }
        \label{fig: ac}
\end{figure}

To demonstrate the effectiveness of PirateNets, let us revisit the Allen-Cahn benchmark we introduced in Section \ref{sec:init_pathology}.  Here we employ a PirateNet with 3 residual blocks (9 hidden layers) and follow the optimal hyper-parameter settings used in JAX-PI \cite{wang2023expert} for training.  The results are shown in Figure \ref{fig: ac}. As shown at the top panel, the model prediction is in excellent agreement with the reference solution,  achieving a relative $L^2$ error of $2.24 \times 10^{-5}$. Moreover, Table \ref{tab: AC} summarizes the performance of various PINNs variants on this benchmark, indicating that our method achieves the best result ever reported in the PINNs literature for this example. Additionally, the bottom panel visualizes the training loss, test error, and the nonlinearity $\alpha$ of each residual block during training. We observe that the PirateNet yields faster convergence compared to a modified MLP backbone. Interestingly, after a rapid change in the early training phase, the learned nonlinearities $\alpha$ stabilize around $\mathcal{O}({10^{-2}})$ and exhibit minimal variation thereafter. This may imply that for this example,  our model efficiently approximates the PDE solution and minimizes the PDE residual with a relatively modest degree of nonlinearity.

\begin{figure}
    \centering
    \includegraphics[width=0.9\textwidth]{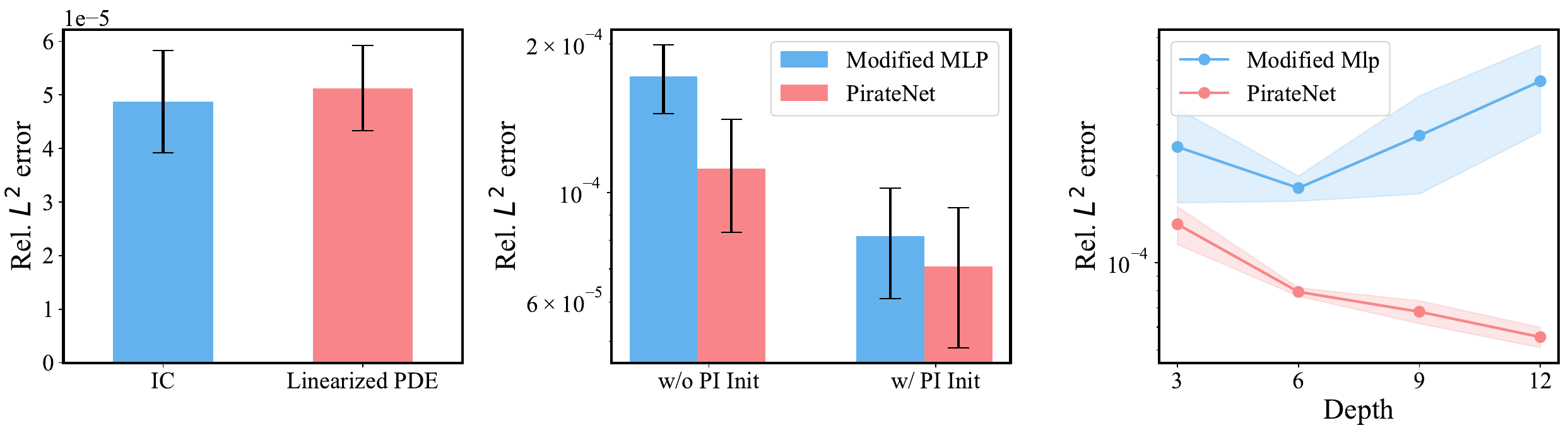}
    \caption{{\em Allen-Cahn equation:} {\em Left:} Relative $L^2$ test errors obtained by a PirateNet with the last layer initialized by the least square solution for fitting the initial condition and the linearized PDE solution, respectively.
    {\em Middle: }  Relative $L^2$ test errors of training a Modified MLP and a PirateNet backbone with or without the physics-informed initalization. Without physics-informed initialization, the final layer defaults to a standard dense layer with weights initialized using the Glorot scheme and biases set to zero.  {\em Right: } Relative $L^2$ errors of training a Modified MLP and a PirateNet backbone of different depth. Each ablation study is performed under the same hyper-parameter settings, with results averaged over 5 random seeds.}
    \label{fig:ac_sweep_2}
\end{figure}

Furthermore, we perform a series of ablation studies to further investigate the performance of PirateNets. First, we explore the impact of various data sources used for the physics-informed initialization of the PirateNet. We examine two different approaches: (a) fitting the initial conditions, and (b) fitting the solution of the linearized PDE:
\begin{align*}
&u_{t}-0.0001 u_{xx} = 0\,, \quad t \in (0,1), x \in (-1,1)\,, \\
&u(0, x) = x^{2} \cos (\pi x)\,,
\end{align*}
under periodic boundary conditions. We evaluate the test error resulting from using these two approaches for the initialization of the last linear layer of the PirateNet. The findings, presented on the left panel of Figure \ref{fig:ac_sweep_2}, indicate comparable performance of both cases.  This result aligns with our expectations, given the negligible discrepancy in the solution of the aforementioned linearized PDE compared to the initial condition across the time domain, due to the small diffusivity of $0.0001$.

Second, we study the influence of the physics-informed initialization across different network architectures. As mentioned in Section \ref{sec: pirate}, any network with a linear final layer can be viewed as a linear combination of basis functions, and is therefore amenable to our proposed physics-informed initialization technique. In this context, we compared the Modified MLP and PirateNet and present the results in the middle panel of Figure \ref{fig:ac_sweep_2}.
It can be observed that the proposed physics-informed initialization can improve the performance of both architectures and notably, the PirateNet outperforms the Modified MLP baseline, irrespective of the initialization method.

Third, we evaluate the effect of network depth on predictive accuracy for different architectures. The results are presented on the right panel of Figure \ref{fig:ac_sweep_2}, revealing a clear difference between Modified MLP and PirateNet.  Contrary to the increasing error observed in deeper Modified MLP networks, PirateNets demonstrate a consistent improvement in accuracy as the network depth increases.  This trend not only underscores PirateNet's superior performance but also highlights its enhanced scalability and robustness.

\begin{table}
    \renewcommand{\arraystretch}{1.4}
    \centering
    \begin{tabular}{l|c}
    \hline
    \textbf{Method}   & \textbf{Relative $L^2$ error}  \\
     \hline
      Original formulation of Raissi {\it et al.} \cite{raissi2019physics}    &  $4.98 \times 10^{-1}$ \\
      Adaptive time sampling \cite{wight2020solving} & $2.33 \times 10^{-2}$ \\
       Self-attention \cite{mcclenny2020self} & $2.10 \times 10^{-2}$  \\
       Time marching \cite{mattey2022novel}  & $1.68 \times 10^{-2}$ \\
       Causal training \cite{wang2022respecting} & $1.39 \times 10^{-4}$ \\
       Dirac delta function causal training \cite{es2023optimal}  & $6.29  \times 10^{-5}$ \\
       JAX-PI \cite{wang2023expert} & $5.37 \times 10^{-5}$ \\
        RBA-PINNs  \cite{anagnostopoulos2023residual}  & $4.55 \times 10^{-5}$ \\
       \textbf{PirateNet (Ours)} & $\mathbf{2.24 \times 10^{-5}}$ \\
    \hline
    \end{tabular}
    \caption{{\em Allen-Cahn equation:} Relative $L^2$ test errors obtained by different PINNs variants.}
    \label{tab: AC}
\end{table}


\subsection{Korteweg–De Vries equation}

Next, we explore the one-dimensional Korteweg–De Vries (KdV) equation, a fundamental model used to describe the dynamics of solitary waves, or solitons. The KdV equation is expressed as follows:
\begin{align*}
& u_t + \eta u u_x + \mu^2 u_{x x x} = 0\,, \quad t \in(0,1), \quad x \in(-1,1)\,, \\
& u(x, 0) = \cos (\pi x)\,, \\
& u(t,-1) = u(t, 1)\,,
\end{align*}
where $\eta$ governs the strength of the nonlinearity, while $\mu$ controls the dispersion level. Under the KdV dynamics,  this initial wave evolves into a series of solitary-type waves. For our study, we adopt the classical parameters of the KdV equation, setting $\eta = 1$ and $\mu = 0.022$ \cite{zabusky1965interaction}.

\begin{figure}
     \centering
     \begin{subfigure}[b]{0.9\textwidth}
         \centering
         \includegraphics[width=\textwidth]{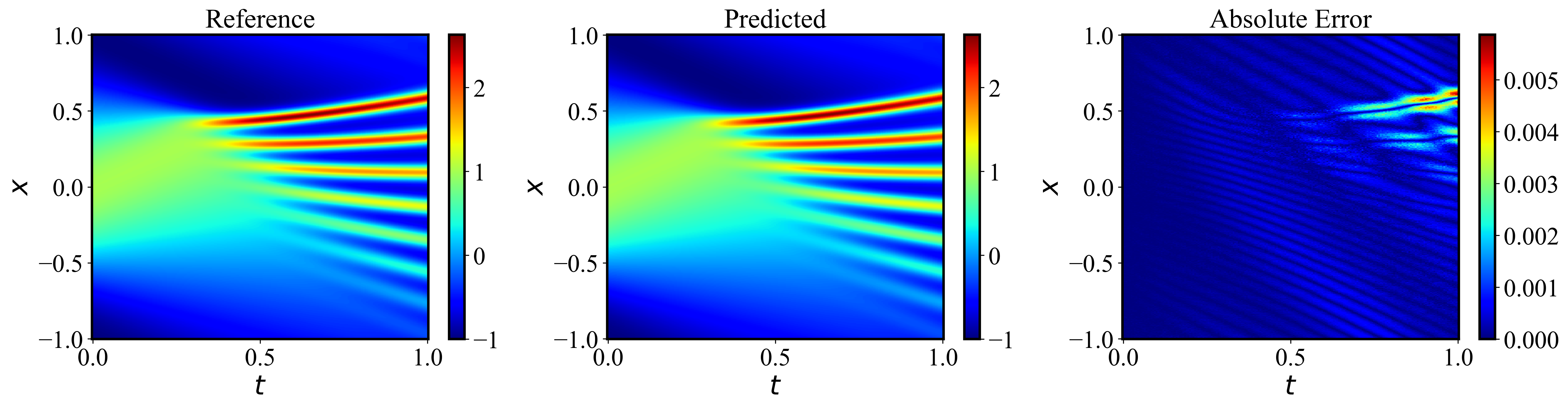}
         \label{fig:kdv_pred}
     \end{subfigure}
     \hfill
     \begin{subfigure}[b]{0.9\textwidth}
         \centering
         \includegraphics[width=\textwidth]{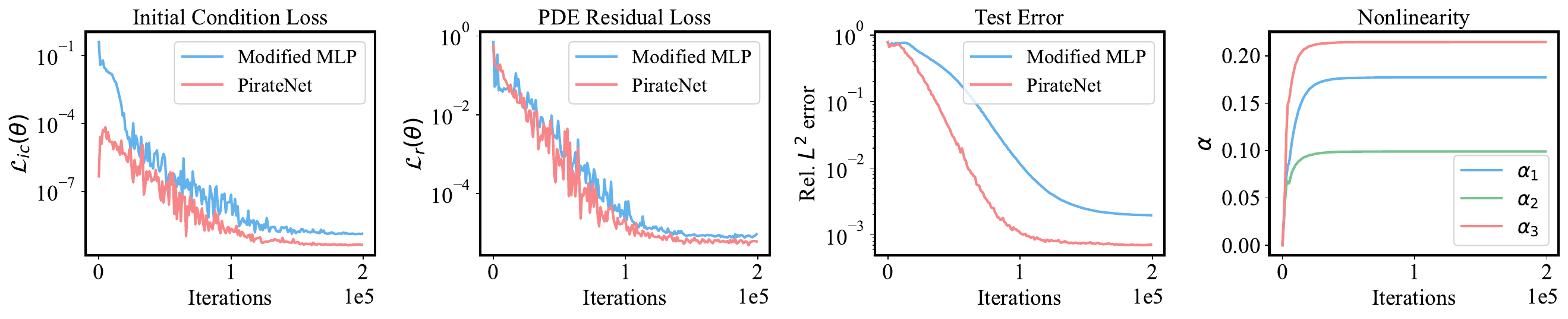}
         \label{fig:kdv_pirate_stat}
     \end{subfigure}
        \caption{{\em Korteweg–De Vries equation:} {\em Top:} Comparison between the solution predicted by a trained PirateNet and the reference solution. The detailed hyper-parameter settings are presented in Table \ref{tab: kdv_config}.  {\em Bottom:}   Convergence of the initial condition loss, the PDE residual loss, and the relative $L^2$ error during the training of a PirateNet and a Modified MLP backbone, alongside the evolution of nonlinearities in each PirateNet residual block. }
        \label{fig:kdv}
\end{figure}

We employ a PirateNet with 3 residual blocks (9 hidden layers) and train the model following the training pipeline and recommended hyper-parameter settings used in JAX-PI \cite{wang2023expert}.  The top panel of Figure \ref{fig:kdv} showcases the predicted solution from the trained model which exhibits an excellent agreement with the numerical solution,  resulting in a relative $L^2$ test error of  $4.27 \times 10^{-4}$.  
For further context, in Table \ref{tab: KDV} we provide a comparison of test errors obtained from competing PINNs approaches on the same benchmark. Impressively, our method outperforms the current state-of-the-art \cite{es2023optimal} by approximately an order of magnitude.

Moreover, in the bottom panel of Figure \ref{fig:kdv} we record the loss and test error obtained by a PirateNet and a Modified MLP backbone, as well as the nonlinearity parameter ($\alpha$) attained in each PirateNet residual block during training. 
Consistent with the trends seen in our first example, the PirateNet yields a faster convergence in terms of both loss and test errors compared to modified MLPs. Intriguingly, in this case, the learned nonlinearities ($\alpha$) reach an order of magnitude around 0.1, in contrast to only 0.01 in the first example.  This indicates that the problem is inherently more nonlinear and requires a correspondingly more complex nonlinear mapping for accurate approximation.  

\begin{figure}
    \centering
    \includegraphics[width=0.9\textwidth]{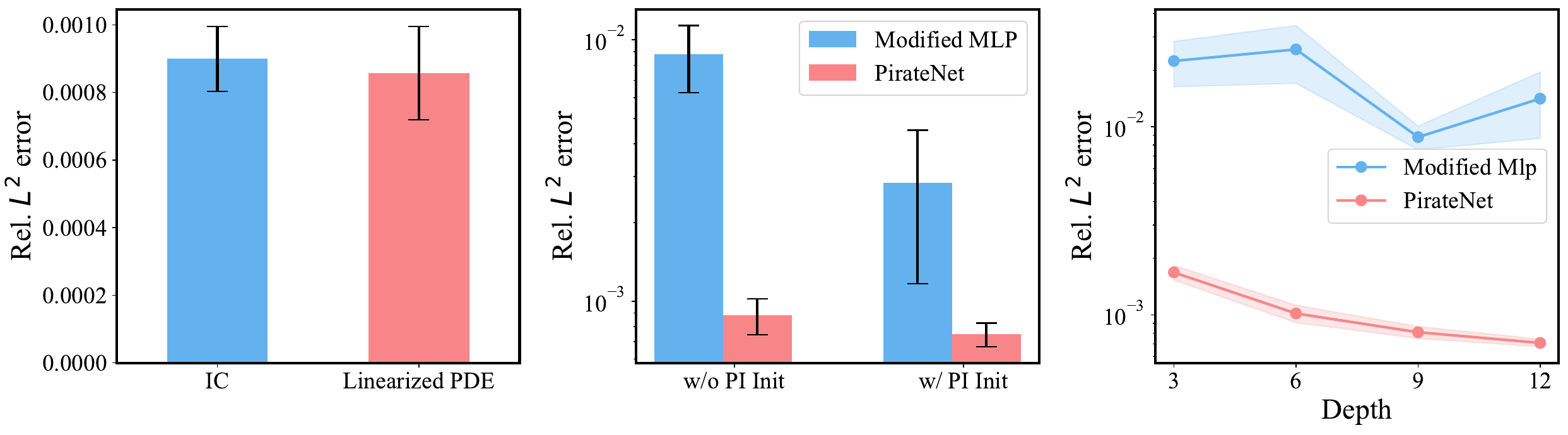}
    \caption{{\em Korteweg–De Vries equation:} {\em Left:} Relative $L^2$ errors of training PirateNet with the last layer initialized by the least square solution to fit the initial condition and the linearized PDE, respectively.
    {\em Middle:}  Relative $L^2$ errors of training a Modified MLP and a PirateNet backbone with or without the physics-informed initialization. Without physics-informed initialization, the final layer defaults to a standard dense layer with weights initialized using the Xavier method and biases set to zero.  {\em Right:} Relative $L^2$ test errors obtained by a Modified MLP and a PirateNet backbone of different depths. Each ablation study is performed under the same hyper-parameter settings, with results averaged over 5 random seeds.}
    \label{fig:kdv_sweep}
\end{figure}

To further validate our conclusions, we conduct the same ablation studies as in our first example. The results are summarized in Figure \ref{fig:kdv_sweep}. The left and the middle panel of the figure reinforce our earlier conclusion that using data from initial conditions or linearized PDEs leads to similar predictive accuracy and the physics-informed initialization can benefit both PirateNet and Modified MLP.
However, it is important to note that PirateNet significantly outperforms Modified MLP for this example, even in the absence of physics-informed initialization.  Finally, the right panel of the figure further affirms the scalability of the PirateNet architecture, as observed in the first example. This evidence highlights the adaptability and efficiency of the PirateNet architecture in tackling complex, nonlinear problems.

\begin{table}
    \renewcommand{\arraystretch}{1.4}
    \centering
    \begin{tabular}{l|c}
    \hline
    \textbf{Method}   & \textbf{Relative $L^2$ error}  \\
     \hline
       Dirac delta function causal training \cite{es2023optimal}  & $2.45 \times 10^{-3}$ \\
       JAX-PI \cite{wang2023expert} & $1.96 \times 10^{-3}$ \\
       \textbf{PirateNet (Ours)} & $\mathbf{4.27 \times 10^{-4}}$ \\
    \hline
    \end{tabular}
    \caption{{\em Korteweg–De Vries equation:} Relative $L^2$ errors obtained by different approaches.}
    \label{tab: KDV}
\end{table}

\subsection{Grey-Scott equation}
\label{sec: gs}



In this example, we solve the 2D Grey-Scott equation, a reaction-diffusion system that describes the interaction of two chemical species. The form of this PDE is given as follows 
\begin{align*}
    u_t &=\epsilon_1 \Delta u + b_1(1-u) - c_1 u v^2,  \quad t \in (0, 2)\,, \ (x, y) \in (-1, 1)^2\,, \\
    v_t &=\epsilon_2 \Delta v - b_2 v + c_2 u v^2\,, \quad t \in (0, 2)\,, \ (x, y) \in (-1, 1)^2\,,
\end{align*}
subject to the  periodic boundary conditions and the initial conditions
\begin{align*}
    &u_0(x, y) = 1 - \exp(-10 ((x + 0.05)^2 + (y + 0.02)^2))\,, \\
    &v_0(x, y) = 1 - \exp(-10 ((x - 0.05)^2 + (y - 0.02)^2))\,.
\end{align*}
Here $u$ and $v$  represent the concentrations of the two species. The system can generate a wide range of patterns, including spots, stripes, and more complex forms, depending on the parameters chosen.  For this example, we set $\epsilon_1 =0.2, \epsilon_2=0.1, b_1=40, b_2=100, c_1=c_2=1,000$, resulting in a beautiful random-seeming ``rolls''.

We employ a standard time-marching strategy to train our PINN models, as outlined in various studies \cite{wight2020solving, krishnapriyan2021characterizing, wang2023expert}. Specifically, we divide the temporal domain $[0, 1]$ into 10 equal intervals, employ the PirateNet architecture as a backbone and train a separate PINN model for each time window.  The initial condition of each segment is given by the predicted solution at the end of its preceding interval. We initialize the last layer of each PirateNet by the least squares fitting of the associated initial conditions.

\begin{figure}
     \centering
     \begin{subfigure}[b]{0.9\textwidth}
         \centering
         \includegraphics[width=\textwidth]{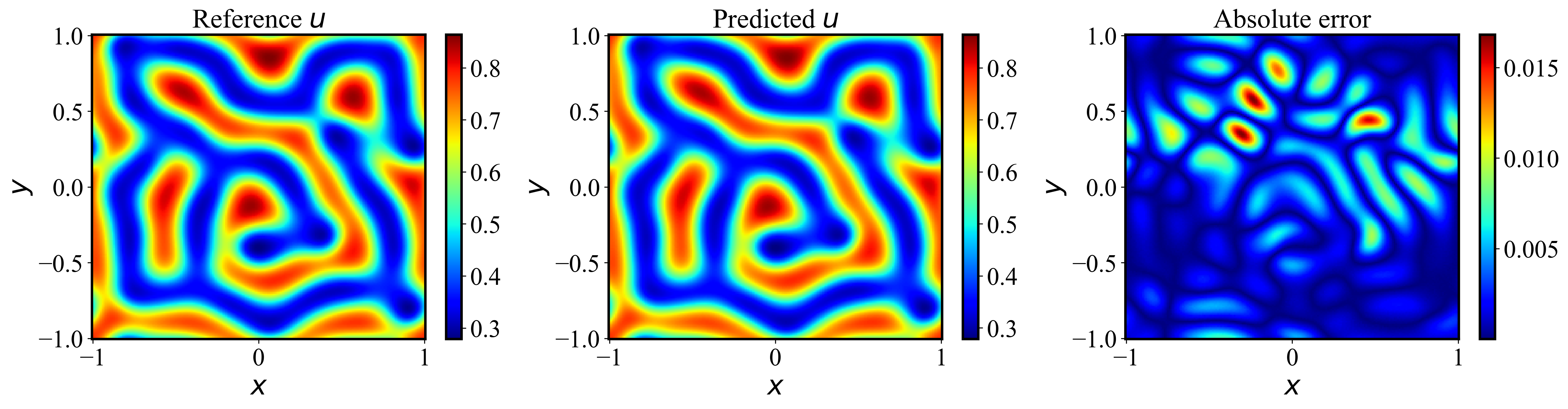}
         \label{fig:gs_u_pred}
     \end{subfigure}
     \hfill
     \begin{subfigure}[b]{0.9\textwidth}
         \centering
         \includegraphics[width=\textwidth]{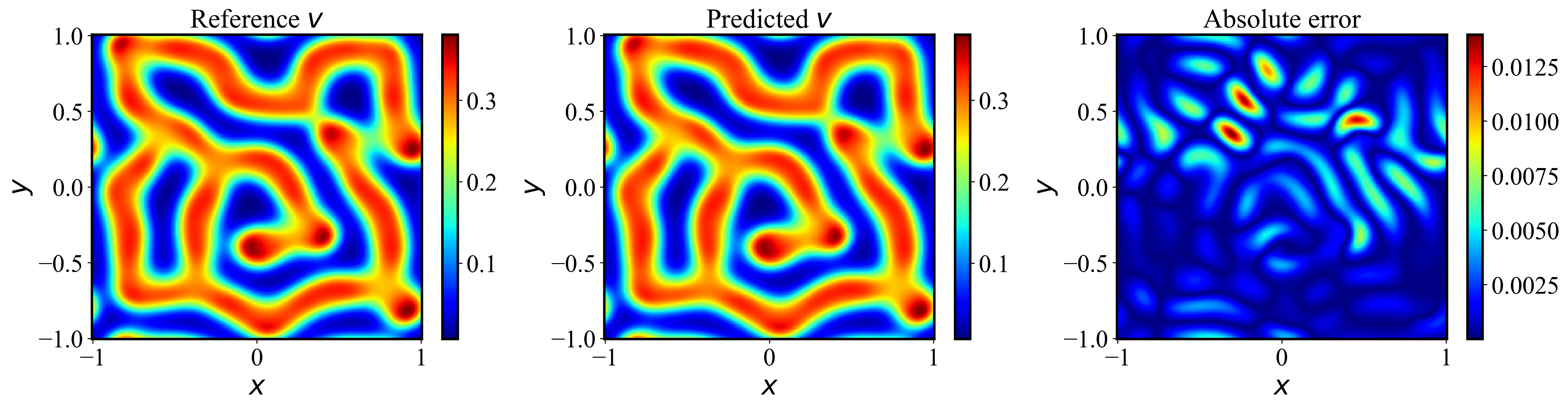}
         \label{fig:gs_v_pred}
     \end{subfigure}
    
        \caption{{\em Grey-Scott equation:} Comparisons between the solutions predicted by a trained PirateNet and the reference solutions. The detailed hyper-parameter settings are presented in Table \ref{tab: gs_config}. }
        \label{fig:gs_pred}
\end{figure}

\begin{figure}
    \centering
    \includegraphics[width=0.9\textwidth]{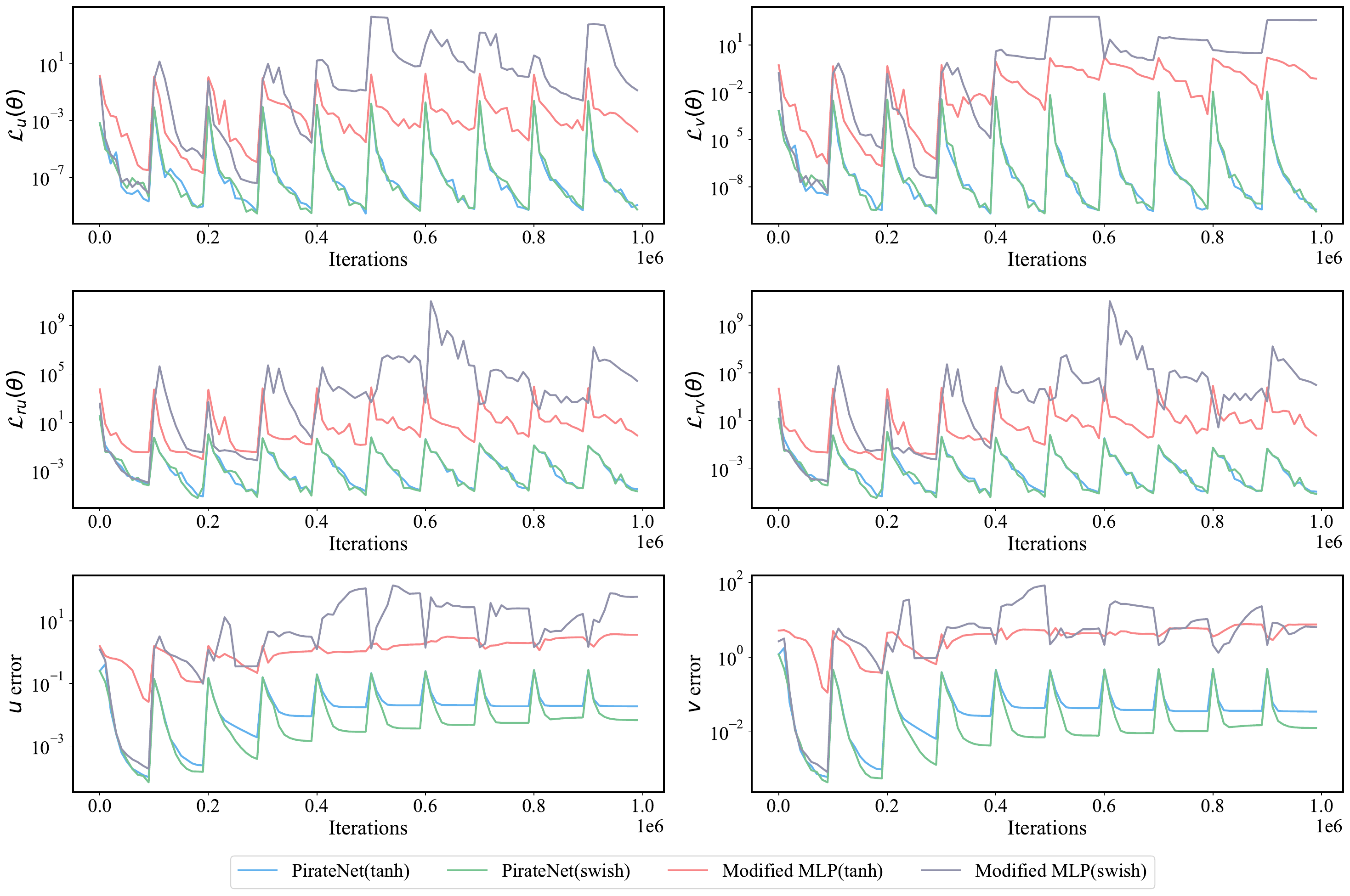}
    \caption{{\em Grey-Scott equation:} Convergence of the initial condition losses, the PDE residual losses and the relative $L^2$ errors during the training of PirateNet and Modified MLP backbones with different activation functions.}
    \label{fig:gs_loss}
\end{figure}

Figure \ref{fig:gs_pred} displays the predicted solutions of the PirateNet model at the final time $T=1$. We can see a good alignment between our predictions and the corresponding numerical estimations. The resulting relative $L^2$ errors of $u$ and $v$ are $3.61 \times 10^{-3}$ and $9.39 \times 10^{-3}$, respectively. Moreover, we compare the performance of PirateNet and Modified MLP backbones using different activation functions under the same hyper-parameter settings. As illustrated in Figure \ref{fig:gs_loss}, PirateNets consistently achieve stable loss convergence with both Tanh and swish activations, while the swish activation yields better accuracy for this example. In contrast, the Modified MLP exhibits significant instability in training loss, characterized by frequent spikes and ultimately failing to approximate the PDE solution. This underscores PirateNet's robustness in terms of activation function choice.  The remarkable training stability of PirateNets can be attributed to its initialization as a shallow network, which is inherently easier to train and more resilient to larger learning rates in the early stages of training.

\begin{figure}
    \centering
    \includegraphics[width=0.7\textwidth]{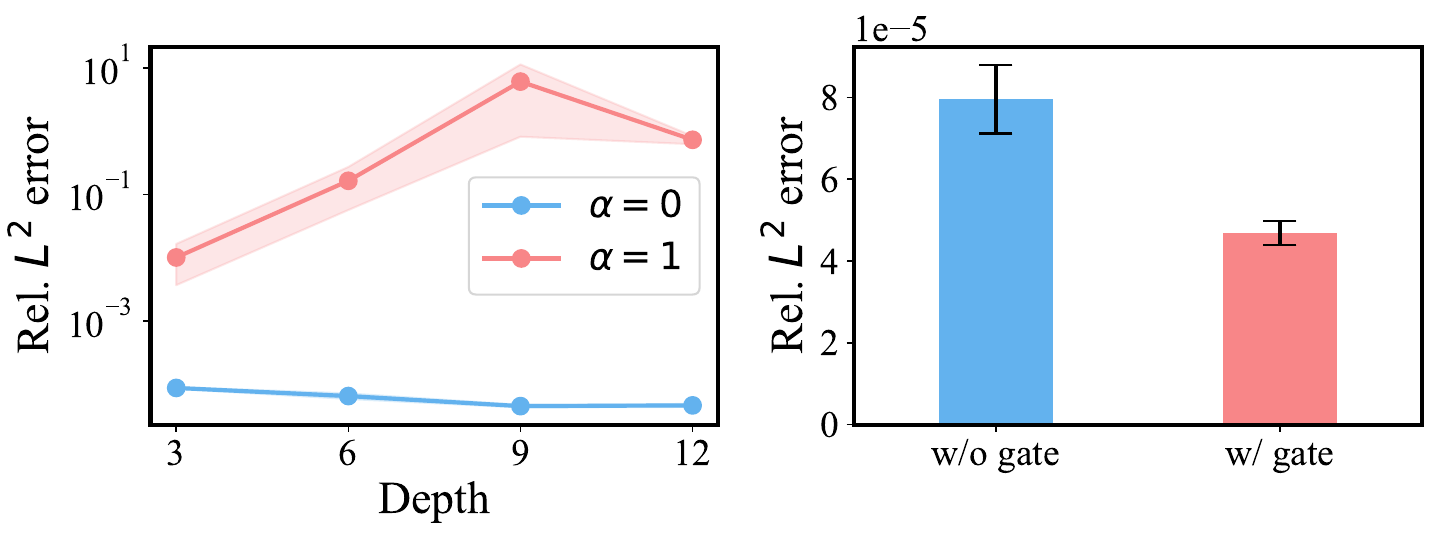}
    \caption{{\em Grey-Scott equation:} {\em Left:} Relative $L^2$ errors of training PirateNet of different depth, with the nonlinearity parameter $\alpha$ of each residual block initialized to 0 or 1. {\em Right:}  Relative $L^2$ errors of training PirateNet with and without gating operations, respectively. 
    Each ablation study is performed under the same hyper-parameter settings. All models are trained for a single time window, with the results averaged over five random seeds.}
    \label{fig:gs_sweep}
\end{figure}

To more thoroughly validate the efficacy of individual components in the PirateNet architecture, we carried out two ablation studies. The goal of the first study is to assess the impact of initializing the nonlinearity parameter $\alpha=0$ of each residual block. For this purpose, we train PirateNet models of varying depths with initialized $\alpha$ at $0$ or $1$, respectively, to solve the PDE system within a time window $[0, 0.1]$.
The results are summarized on the left panel of Figure \ref{fig:gs_sweep}, which clearly demonstrate that initializing $\alpha=0$  significantly enhances accuracy by one or two orders of magnitude.   Conversely, initializing $\alpha=1$ results in an increased error, particularly when training deeper networks. 

The second ablation study aims to examine the role of gating operations, as described in Equations \eqref{eq: gate1} and \eqref{eq: gate2}. To this end, we fix a network depth of 9 and train the PirateNet with and without gating operations over 5 random seeds. As presented on the right panel of Figure \ref{fig:gs_sweep}, we can see that the incorporation of gating operations further reduces predictive errors. These findings strongly support the chosen initialization of $\alpha=0$ and the design of the gating operation as key factors in the successful training of deep PirateNet models.

\subsection{Ginzburg-Landau equation}

In this example, we aim to solve a  scalar PDE involving a complex variable. To this end, we consider the complex Ginzburg-Landau equation in 2D of the form
\begin{align*}
    \frac{\partial A}{\partial t}= \epsilon \Delta A + \mu A - \gamma  A|A|^2\,, \quad t \in (0, 1)\,,
    \ (x, y) \in (-1, 1)^2\,,
\end{align*}
with an initial condition
\begin{align*}
    A_0(x, y) = (10y + 10 i  x) \exp\left(-0.01 (2500 x^2 + 2500 y^2)\right)\,,
\end{align*}
where $A$ denotes a complex solution and we set $\epsilon=0.004$, $\mu = 10$ and $\gamma= 10 + 15i$.
It describes a vast variety of phenomena from nonlinear waves to second-order phase transitions, from superconductivity, superfluidity, and Bose-Einstein condensation, to liquid crystals and strings in field theory.

By denoting $A = u + i v$, we can decompose the equation into real and imaginary components, resulting in the following system of PDEs,
\begin{align*}
     \frac{\partial u}{\partial t} &= \epsilon \Delta u + \mu (u - (u-1.5 v) (u^2 + v^2))\,, \\
      \frac{\partial v}{\partial t} &= \epsilon \Delta v + \mu (v - (v + 1.5 u) (u^2 + v^2))\,.
\end{align*}
In our experiments, we solve the decoupled equations using PINNs with a time-marching strategy, analogous to the approach detailed in Section \ref{sec: gs}. Specifically, we partition the time domain into five equal intervals, each of length 0.2, and train a PINN for each interval in sequence. The initial condition of each interval is given by the predicted solution at the end of the previous interval.  For PirateNets, we initialize its final layer to align with these initial conditions via least squares.

\begin{figure}
     \centering
     \begin{subfigure}[b]{0.9\textwidth}
         \centering
         \includegraphics[width=\textwidth]{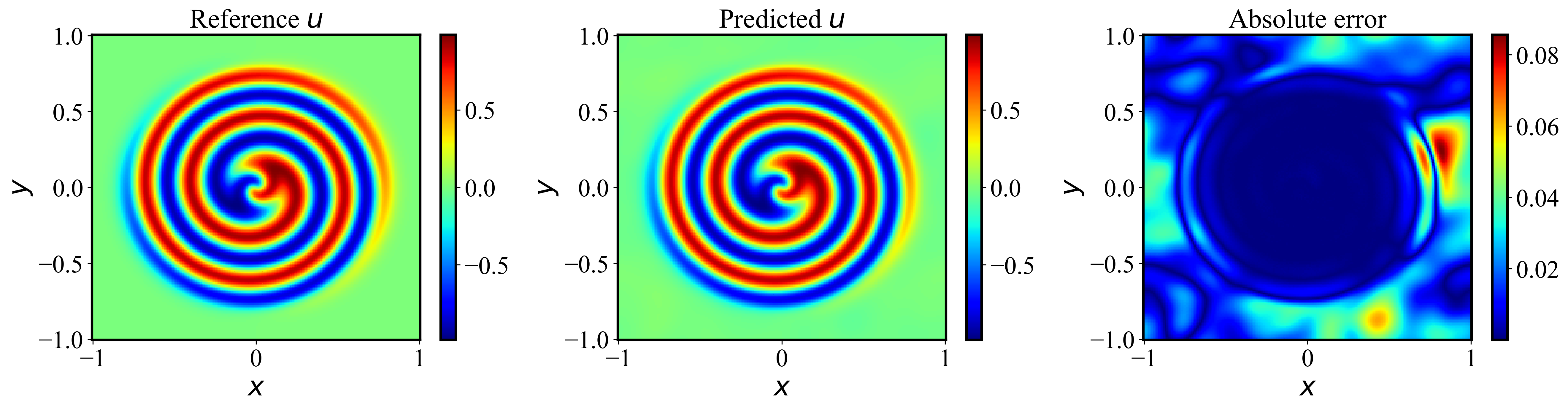}
         \label{fig:gl_u_pred}
     \end{subfigure}
     \hfill
     \begin{subfigure}[b]{0.9\textwidth}
         \centering
         \includegraphics[width=\textwidth]{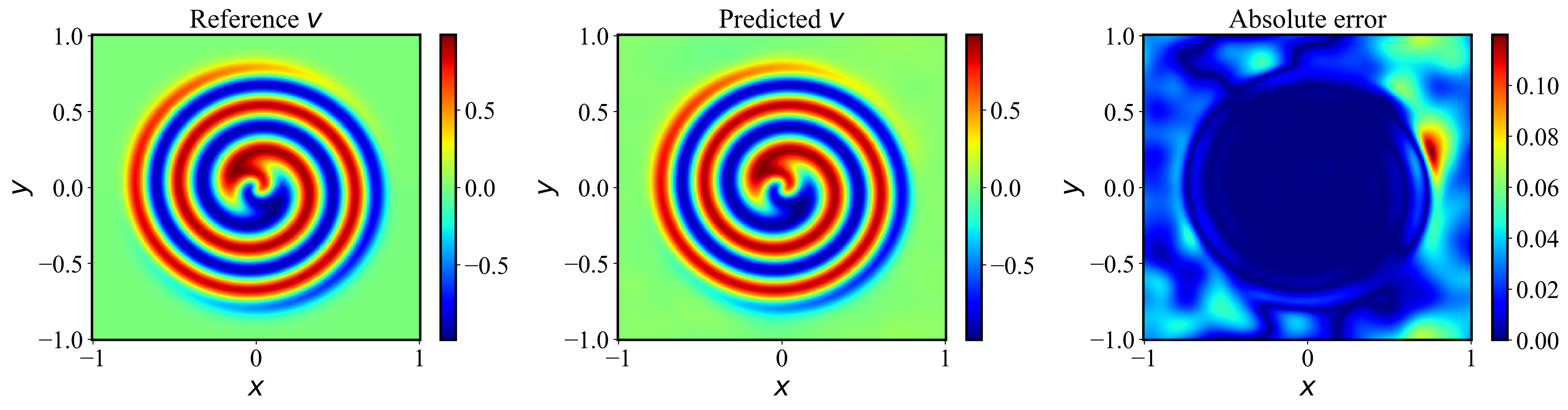}
         \label{fig:gl_v_pred}
     \end{subfigure}
        \caption{{\em Ginzburg-Landau equation:} Comparisons between the solutions predicted by a trained PirateNet and the reference solutions.  The detailed hyper-parameter settings are presented in Table \ref{tab: gl_config}.}
        \label{fig:gl_pred}
\end{figure}

\begin{figure}
    \centering
    \includegraphics[width=0.9\textwidth]{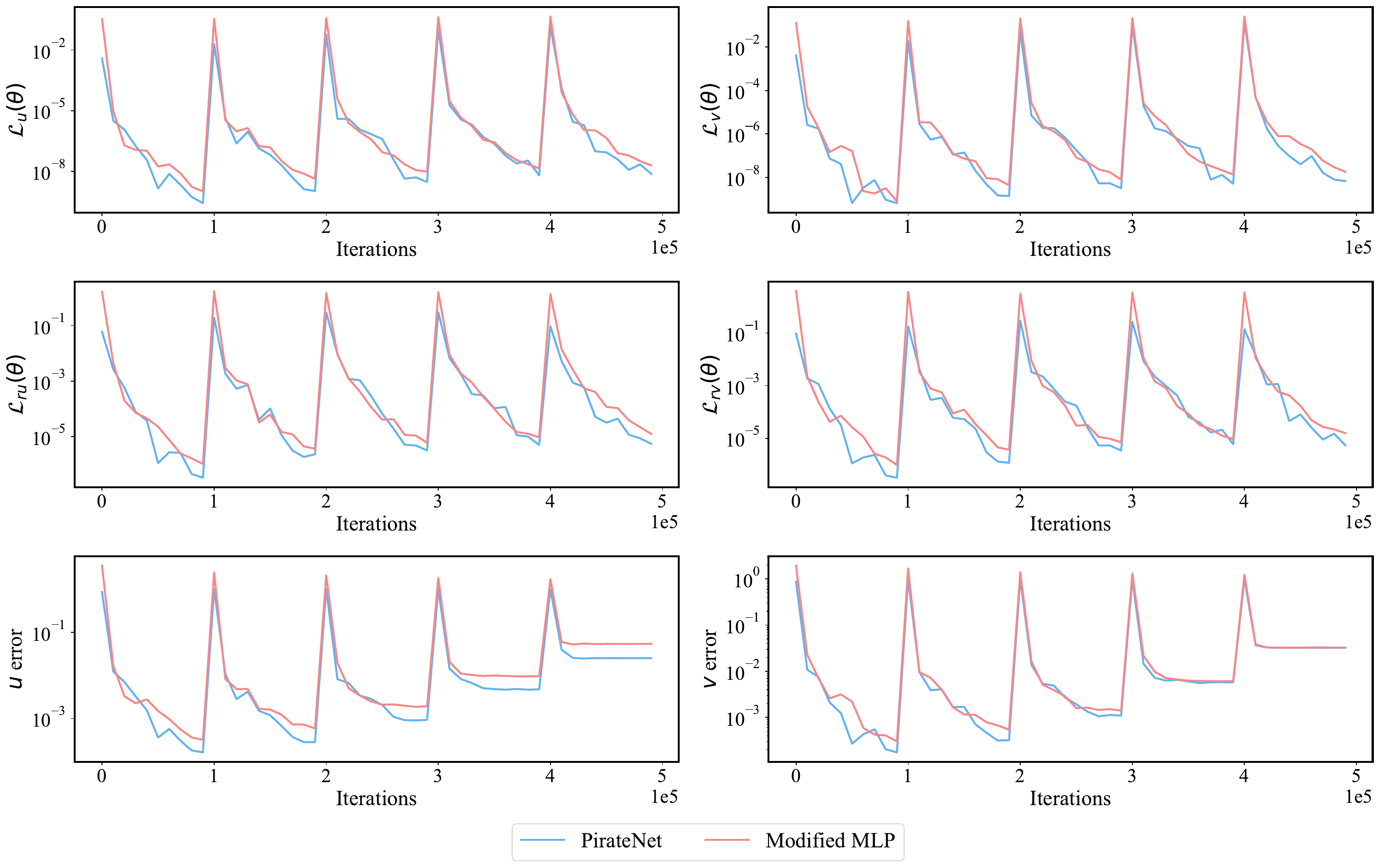}
    \caption{{\em Ginzburg-Landau equation:} Convergence of the initial condition losses, the PDE residual losses, and the relative $L^2$ errors during the training of a PirateNet and a Modified MLP backbone with different activation functions.}
    \label{fig:gl_loss}
\end{figure}

Figure \ref{fig:gl_pred} displays the predicted real and imaginary parts of the solution obtained by using the PirateNet backbone. The predictions show a good agreement with the corresponding ground truth. The computed $L^2$ errors for the real part $u$ and the imaginary part $v$ are $1.49 \times 10^{-2}$ and $1.90 \times 10^{-2}$, respectively. These results outperform the accuracy achieved by the Modified MLP backbone, which recorded errors of $3.20 \times 10^{-2}$ for $u$ and $1.94 \times 10^{-2}$ for $v$.  This is further reflected in the loss convergence patterns observed in Figure \ref{fig:gl_loss}.

\begin{figure}
    \centering
    \includegraphics[width=0.7\textwidth]{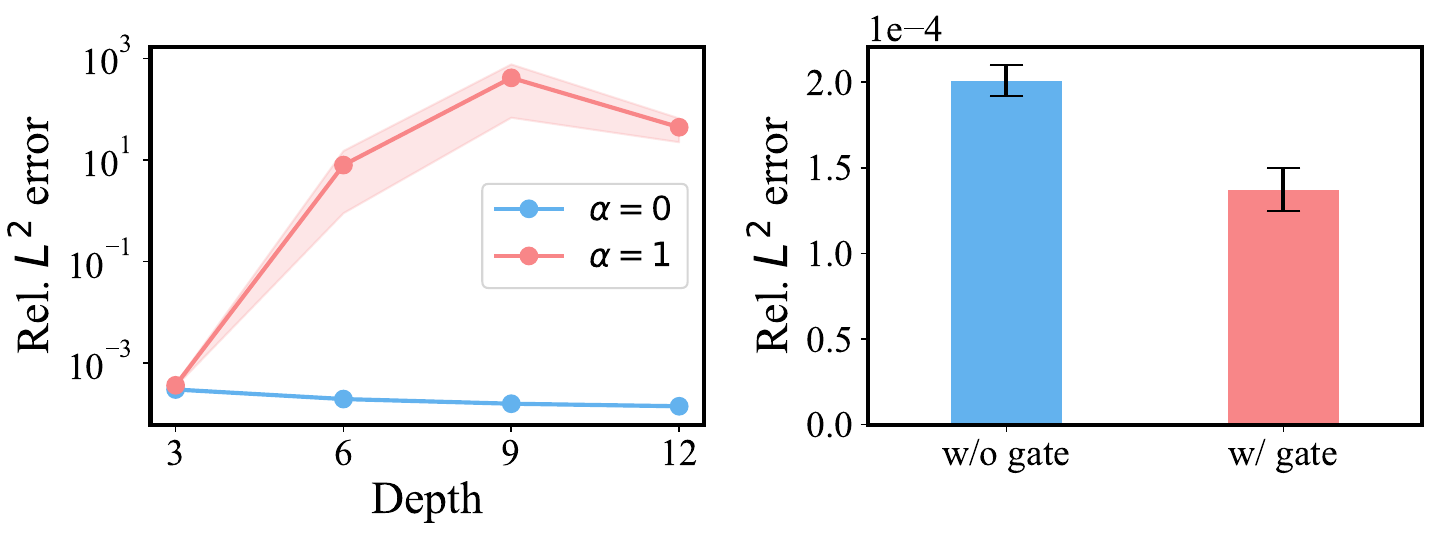}
    \caption{{\em Ginzburg-Landau equation:} {\em Left:} Relative $L^2$ errors of training PirateNet of different depth, with the nonlinearity parameter $\alpha$ of each residual block initialized to 0 or 1. {\em Right:}  Relative $L^2$ errors of training PirateNet with and without gating operations, respectively. 
    Each ablation study is performed under the same hyper-parameter setting. All models are trained within the first time window, with the results averaged over five random seeds.}
    \label{fig:gl_sweep}
\end{figure}

Furthermore,  we perform the same ablation studies as in Section \ref{sec: gs} to further affirm the performance contributions of each element in the PirateNet architecture.  As illustrated in the left panel of Figure \ref{fig:gl_sweep}, setting $\alpha=1$ at initialization results in complete model collapse for deeper networks,  while setting $\alpha=0$ ensures stable training and consistently improved accuracy as network depth increases.  In addition,  the right panel of the figure supports our earlier finding that the use of gating operations enhances the model's accuracy.

\subsection{Lid-driven Cavity flow}

In this example, we focus on a classical benchmark problem in computational fluid dynamics that involves simulating the movement of an incompressible fluid within a two-dimensional square cavity.
The system is governed by the incompressible Navier–Stokes equations written in a non-dimensional form
\begin{align*}
    \mathbf{u} \cdot \nabla \mathbf{u}+\nabla p-\frac{1}{R e} \Delta \mathbf{u}&=0\,, \quad  (x,y) \in (0,1)^2\,, \\
    \nabla \cdot \mathbf{u}&=0\,, \quad  (x,y) \in (0,1)^2\,,
\end{align*}
where $\mathbf{u} = (u, v)$ denotes the velocity in $x$ and $y$ directions, respectively, and $p$ is the scalar pressure field.  To avoid discontinuities of the top lid boundary conditions at two corners, we reformulate the  boundary condition  as follows
\begin{align}
& u(x, y)=1-\frac{\cosh \left(C_0(x-0.5)\right)}{\cosh \left(0.5 C_0\right)}\,, \quad v(x, y)=0\,,
\end{align}
where $x \in [0, 1], y=1, C_0 = 50$. For the other three walls, we enforce a no-slip boundary condition. Our goal is to obtain the velocity and pressure field corresponding to a Reynolds number of $3200$.

As illustrated by recent literature \cite{wang2023expert,wang2023solution,cao2023tsonn}, training PINNs directly at high Reynolds numbers often leads to instability and a tendency to converge to erroneous solutions. To address this, we employ a curriculum training strategy \cite{krishnapriyan2021characterizing,wang2023expert}, initially training  PINNs at a lower Reynolds number, and gradually increasing the Reynolds numbers during training. This approach allows the model parameters optimized at lower Reynolds numbers to serve as a favorable initialization for training at higher Reynolds numbers.
Specifically, we train PINNs across a progressively increasing sequence of Reynolds numbers: $[100, 400, 1000, 1600, 3200]$. For each Re, we conduct training the model for $ 10^4$, $2 \times 10^4$, $5 \times 10^4$, $5 \times 10^4$, and $5 \times 10^5$ iterations, respectively.

\begin{figure}
    \centering
   \includegraphics[width=0.9\textwidth]{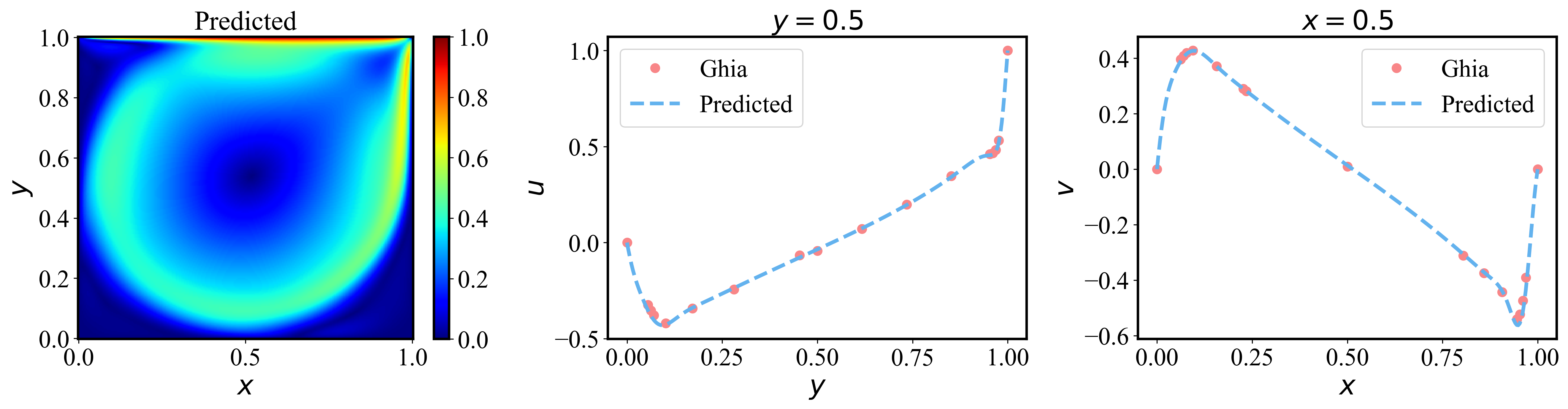}
    \caption{{\em Lid-driven cavity (Re=3200):} {\em Left:} Predicted velocity of the fine-tuned model. {\em Right:} Comparison of the predicted velocity profiles on the vertical and horizontal center-lines against Ghia {\em et al.} \cite{GHIA1982387}. The resulting relative $L^2$ error against the reference solution is $4.21 \times 10^{-2}$.
    The detailed hyper-parameter settings are presented in Table \ref{tab: ldc_config}. }
    \label{fig:ldc_pred}
\end{figure}

\begin{figure}
    \centering
   \includegraphics[width=0.8\textwidth]{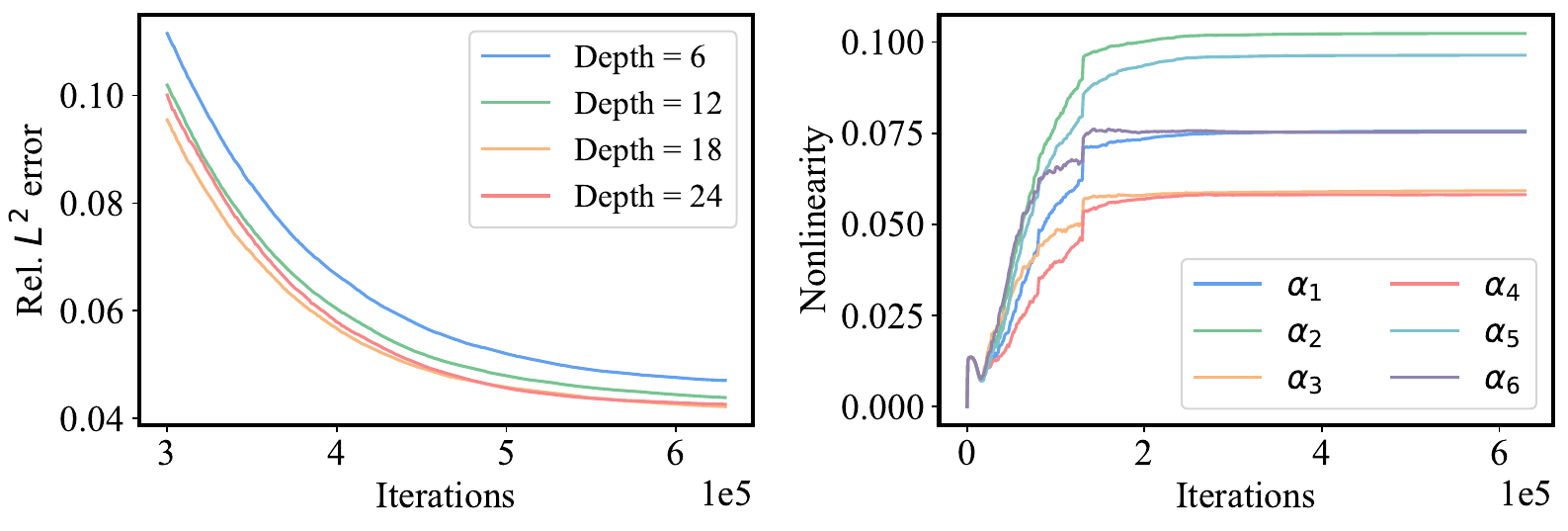}
    \caption{{\em Lid-driven cavity (Re=$3200$):} {\em Left:} Convergence of the relative $L^2$ errors during the training of PirateNets of different depths for the final training stage at $Re=3200$. All experiments are performed under the same hyper-parameter setting.
      {\em Right:} Evolution of nonlinearities in each PirateNet residual block of training a PirateNet with 6 residual blocks (18 hidden layers). }
    \label{fig:ldc_error_alpha}
\end{figure}

Figure \ref{fig:ldc_pred} plots the predicted velocity field for a Reynolds number of $Re=3200$.  The predictions align closely with the reference results from Ghia {\em et al.} \cite{GHIA1982387}. 
This is evidenced by a relatively low $L^2$ error of $4.21 \times 10^{-2}$, a significant improvement over the $1.58 \times 10^{-1}$ error reported by JAXPI \cite{wang2023expert}. Our ablation study on the model's depth is visualized on the right panel of Figure \ref{fig:ldc_error_alpha}, which reveals consistent performance enhancements as the depth increases within a moderately large range.
However,  the model seems to be saturated at depths greater than 20,  yet the training remains stable and produces results similar to the one at a depth of 18. 
This observation highlights PirateNet's impressive scalability and stability in handling complex, highly nonlinear problems and motivates future advancements in the model architecture. Interestingly, we also examine the evolution of the nonlinearities $\alpha$  in each residual block of the network. A noticeable increase in $\alpha$ is observed when the model transitions to training at higher Reynolds numbers.  This trend suggests that the model requires a higher degree of nonlinearity in order to minimize the associated PDE residuals at higher Reynolds numbers.

\section{Discussion}
\label{sec: discussion}

We introduce PirateNets, a new backbone for efficient training of deep PINN models. We provide theoretical arguments and illustrative numerical experiments that demonstrate the instability of training PINNs when using deep MLPs. To explore the underlying reasons behind it, we propose a fundamental hypothesis that a good convergence in the PINNs training loss implies a good convergence of the PDE solution in $C^k$, depending on the order of the PDE system. Furthermore, under mild conditions, we rigorously prove that this hypothesis holds for linear elliptic and parabolic PDEs.  This understanding allows us to shift our focus from analyzing the network's PDE residuals to examining the MLP derivatives, both at initialization and during training. Our theoretical and empirical analysis reveals that standard initialization schemes result in pathological initialization of MLP derivatives, adversely affecting the trainability and robustness of PINNs.

Motivated by these insights, we introduce PirateNets; a novel architecture with adaptive residual connections. This design choice ensures that the network is initially a linear combination of a chosen basis, effectively addressing the deep MLP derivatives' initialization issues.  It is noteworthy to observe that PirateNet's approximation capacity is gradually recovered during training, thanks to the trainable parameter $\alpha$ in each residual block.  Additionally, the initial linearity of the model offers a unique advantage: it can be tailored to any kind of existing dataset by initializing the final network layer using solutions derived from corresponding least squares problems. Taking all together,  our developments provide new insights into training PINNs and yield the state-of-art accuracy across a wide range of benchmarks in computational physics.

While this study marks a substantial advancement in understanding the training of deep physics-informed neural networks, we have to admit that training PINNs still requires significant computational resources and tends to be time-consuming. Future efforts should concentrate on accelerating this process. One promising direction could involve optimizing PirateNet's coordinate embeddings and tailoring them to specific PDEs. Besides, the physics-informed initialization introduced here opens up new avenues for integrating physical principles into machine learning models. It would be interesting to extend this idea further to the field of operator learning. This may motivate the more efficient design of neural operators in solving parametric PDEs. We believe that pursuing these lines of exploration is crucial, as they are likely to form a key component in evolving physics-informed machine learning into a powerful and reliable tool for computational science and engineering.


\section*{Acknowledgments}
We would like to acknowledge support from the US Department of Energy under the Advanced Scientific Computing Research program (grant DE-SC0024563). We also thank the developers of the software that enabled our research, including JAX \cite{jax2018github}, JAX-CFD\cite{Kochkov2021}, Matplotlib \cite{hunter2007matplotlib}, and NumPy \cite{harris2020array}.

\bibliographystyle{unsrt}  
\bibliography{references}

\appendix

\section{Proofs}\label{app:a}

\begin{proof}[Proof of Proposition \ref{prop: elliptic}]

Note that the PINN solution $u_\theta \in C^\infty(\Omega)$ if the activation function is smooth. We define the interior residual $\mc{R}_{\rm int, \theta} = \mc{D}[u_\theta] - f$ and the spatial boundary residual $\mc{R}_{\rm bc, \theta} = u_\theta$. With this notation, we see
\begin{align*}
    \h{\mc{L}}(\theta) =  \norm{\mc{R}_{\rm bc, \theta}}_{L^2(\p \Omega)}^2 + \norm{\mc{R}_{\rm int, \theta}}_{L^2(\Omega)}^2\,.
\end{align*}
Let $u^* = u_{\theta^*}$ be a PINN solution generated by the algorithm. It is easy to see that the error $\h{u} = u - u^*$ satisfies $\mc{D}[\h{u}] = -\mc{R}_{\rm int, \theta^*}$ with $\h{u} = \mc{R}_{\rm bc, \theta^*}$ on $\p \Omega$. By interior regularity \cite[Section 6, Theorem 2]{evans2022partial}, we have for any $m \ge 0$, there holds $\h{u} \in H_{\rm loc}^{m + 2}(\Omega)$ with for $V  \subset  \subset \Omega$, 
\begin{align*}
    \norm{\h{u}}_{H^{m + 2}(V)} \le C (\norm{\mc{R}_{\rm int, \theta^*}}_{H^m(\Omega)} + \norm{\h{u}}_{L^2(\Omega)})\,.  
\end{align*}
Let $\mc{M}$ be the Dirichlet map defined by $u := \mc{M}[g]$ if and only if $\mc{D}[u] = 0$ with $u = g$ on $\p \Omega$. It is known that $\mc{M}$ is a bounded linear map from $H^s(\p \Omega)$ to $H^{s + 1/2}(\Omega)$ for any real $s$ \cite{bramble1970rayleigh}. Then, since $\mc{R}_{\rm bc, \theta^*}$ is smooth, we can find a smooth solution such that $\mc{D}[v] = 0$ with $v = \mc{R}_{\rm bc, \theta^*}$ on $\p \Omega$ and there holds 
\begin{equation*}
    \norm{v}_{L^2(\Omega)} \le C \norm{\mc{R}_{\rm bc, \theta^*}}_{L^2(\p \Omega)}\,,
\end{equation*}
which gives $\mc{D}[\h{u} - v] =  \mc{R}_{\rm int, \theta^*}$ with Dirichlet boundary. By the well-posedness, it follows that 
\begin{equation*}
    \norm{\h{u} - v}_{L^2(\Omega)} \le C \norm{\mc{R}_{\rm int, \theta^*}}_{L^2(\Omega)}\,.
\end{equation*}
Hence, we have 
\begin{equation*}
     \norm{\h{u}}_{H^{2}(V)} \le C ( \norm{\mc{R}_{\rm bc, \theta^*}}_{L^2(\p \Omega)} + \norm{\mc{R}_{\rm int, \theta^*}}_{L^2(\Omega)}) \leq C \mc{L}(\theta)^{1/2}\,. \qedhere
\end{equation*} 
\end{proof}

\begin{proof}[Proof of Proposition \ref{prop: parabolic}]
Similarly, we denote the PINN solution by $u^* = u_{\theta^*}$. Then, $\h{u} = u - u^*$ satisfies
\begin{equation*}
    \h{u}_t + \mc{D}[\h{u}] =: \mc{R}_{\rm int, \theta^*}\,,
\end{equation*}
with $\h{u} =: \mc{R}_{\rm bc, \theta^*}$ on $ [0,T] \times \p \Omega  $ and $\h{u} =: \mc{R}_{\rm ic, \theta^*}$ on $\{t = 0\} \times \Omega  $. For a set $Q_r(t, \bb{x}) \subset [0,T] \times \Omega$, the interior estimate of parabolic equation gives \cite[Theorem 2.4.7]{krylov1996lectures}
\begin{align*}
    \norm{\h{u}}_{W^{1,k+2}_2(Q_r)} \le C ( \norm{\mc{R}_{\rm int, \theta^*}}_{L^2([0,T]; L^2(\Omega))} + \norm{\h{u}}_{L^2([0,T]; L^2(\Omega))})\,.
\end{align*}
Let $v$ be the solution to $v_t + \mc{D}[v] = 0$ with $v = \mc{R}_{\rm bc, \theta^*}$ on $ [0,T] \times \p \Omega $ and $v = \mc{R}_{\rm ic, \theta^*}$ on $\{t = 0\} \times \Omega  $. The regularity estimate for $v$ can be given by the standard semigroup method, which we recall below for the reader's convenience. 
We denote by $S(t)$ the strongly continuous analytic semigroup generated by $- \mc{D}$ and recall the Dirichlet map $\mc{M}$ introduced above. 
Then, we have the following representation for the solution $v$ \cite[Section 2]{lasiecka1986galerkin}
\begin{align*}
    v = S(t)[\mc{R}_{\rm ic, \theta^*}] + A  [\mc{R}_{\rm bc, \theta^*}]\,,
\end{align*}
where
\begin{align*}
    A [u] = \mc{D} \int_0^t S(t-z) \mc{M}[u](z) \, \rd z: L^2([0,T]; L^2(\p  \Omega)) \longrightarrow L^2([0,T]; L^2( \Omega))\,.
\end{align*}
It follows that $v \in L^2([0,T]; L^2(\Omega))$ and  
\begin{align*}
    \norm{v}_{L^2([0,T]; L^2(\Omega))}\le C(\norm{\mc{R}_{\rm ic, \theta^*}}_{L^2(\Omega)} + \norm{\mc{R}_{\rm bc, \theta^*}(t)}_{L^2([0,T]; L^2(\p \Omega))})\,.
\end{align*}
Moreover, we have \cite[Section 7, Theorem 5]{evans2022partial}
\begin{align*}
  \norm{\h{u} - v}_{L^2([0,T];H^2(\Omega))} \le C \norm{\mc{R}_{\rm int, \theta^*}}_{L^2([0,T]; L^2(\Omega))}\,.
\end{align*}
Therefore, we have
\begin{align*}
    \|\hat{u}\|_{W_2^{1, k+2}\left(Q_r\right)} \leq C\left(\left\|\mathcal{R}_{\mathrm{int}, \theta^*}\right\|_{L^2\left([0, T] ; L^2(\Omega)\right)}+\left\|\mathcal{R}_{\mathrm{ic}, \theta^*}\right\|_{L^2(\Omega)}+\left\|\mathcal{R}_{\mathrm{bc}, \theta^*}\right\|_{L^2\left([0, T] ; L^2(\partial \Omega)\right)}\right) \leq C \mc{L}(\theta)^{1/2}\,. 
\end{align*}

\end{proof}

\begin{proof}[Proof of Proposition \ref{prop}]
(a) The first statement is a simple consequence of the chain rule. If we further assume the linear regime that $\sigma(\cdot) \approx 1$, then $\operatorname{diag}(\dot{\sigma}(\mathbf{u}_\theta^{(k)}(x))) \approx I$ for all $k$. Then we obtain
            \begin{align}
                \frac{\partial u_\theta}{\partial x}(x) \approx \mathbf{W}^{(L+1)} \cdot \mathbf{W}^{(L)} \cdots \mathbf{W}^{(1)}\,.
            \end{align}
(b) We prove this result by induction on the number of network layers $L$. For $L=1$, the network derivative is given by
\begin{align}
    \frac{\partial u_\theta}{\partial x}(x)=\sum_{k=1}^d W_k^{(2)} \dot{\sigma}\left(W_k^{(1)} x+b^{(1)}\right) W_k^{(1)}\,.
\end{align}
Since $W_k^{(l)}$ is independent from each other for all $l=1,2$ and $k=1,2,\dots, d$, we have
\begin{align} \label{eq: c7}
\operatorname{Var}\left(\frac{\partial u_\theta}{\partial x}(x)\right) & =\sum_{k=1}^d \operatorname{Var}\left(W_k^{(2)} \dot{\sigma}\left(W_k^{(1)} x+b^{(1)}\right) W_k^{(1)}\right)  \notag \\
& =\sum_{k=1}^d \left[ \mathbb{E}\left(W_k^{(2)} \dot{\sigma}\left(W_k^{(1)} x + b^{(1)}\right) W_k^{(1)}\right)^2 - \left(\mathbb{E}\left(W_k^{(2)} \dot{\sigma}\left(W_k^{(1)} x+b^{(1)}\right) W_k^{(1)}\right)\right)^2 \right] \notag\\
& \leq \sum_{k=1}^d \left[ \mathbb{E}\left(W_k^{(2)} W_k^{(1)}\right)^2 - \left(\mathbb{E}\left(W_k^{(2)}\right) \cdot \mathbb{E}\left(\dot{\sigma}\left(W_k^{(1)} x+b^{(1)}\right) W_k^{(1)}\right)\right)^2 \right] \notag \\
& =\sum_{k=1}^d \operatorname{Var}\left(W_k^{(2)}\right) \operatorname{Var}\left(W_k^{(1)}\right) \notag \\
& \lesssim d \cdot \frac{1}{d} \cdot \frac{1}{d}=\frac{1}{d}\,.
\end{align}

Now, assume that this holds for a scalar network of $L$ layers. Then, by (a), for $L+1$, we have
\begin{align}
     \frac{\partial u_\theta}{\partial x}(x) = \prod_{i=L+1}^2\left(\mathbf{W}^{(i)} \cdot \operatorname{diag}\left(\dot{\sigma}\left(\mathbf{u}_\theta^{(i-1)}(x)\right)\right)\right) \cdot \mathbf{W}^{(1)}\,.
\end{align}
Let $X = \prod_{i=L}^2\left(\mathbf{W}^{(i)} \cdot \operatorname{diag}\left(\dot{\sigma}\big(\mathbf{u}_\theta^{(i-1)}(x)\big)\right)\right) \cdot \mathbf{W}^{(1)}$. The key observation is that each element $X_k$ is a scalar network of $L$ layers. By induction, we have $\operatorname{Var}(X_i) \lesssim \frac{1}{d}$ for $k=1,2, \dots, d$.
Therefore,
\begin{align}
     \frac{\partial u_\theta}{\partial x}(x) = \sum_{k=1}^d W^{(L+1)}_k \left.\dot{\sigma}\left(\mathbf{u}_\theta^{(L)}(x)\right)\right) X_k\,.
\end{align}
Similar to \eqref{eq: c7}, we can prove that
\begin{align}
    \frac{\partial u_\theta}{\partial x}(x) \lesssim \frac{1}{d}\,.
\end{align}
Finally, applying Chebyshev's inequality, for any $\epsilon > 0$, we obtain
    \begin{align}
       \operatorname{Var}\left(\frac{\partial u_\theta}{\partial x}(x)\right) \lesssim \frac{1}{d \epsilon^2}\,.
    \end{align}

\end{proof}

\section{Allen-Cahn equation}

\paragraph{Data generation:}  We solve the Allen-Cahn equation using conventional spectral methods. Specifically, assuming periodic boundary conditions, we start from the initial condition $u_0(x) = x^2 \cos(\pi x)$ and integrate the system up to the final time $T=1$. Synthetic validation data is generated using the Chebfun package \cite{driscoll2014chebfun} with a spectral Fourier discretization with 512 modes and a fourth-order stiff time-stepping scheme (ETDRK4) \cite{cox2002exponential} with the time-step size of $10^{-5}$. We record the solution at intervals of 
 $\Delta t = 0.005$, yielding a validation dataset with a resolution of  $200 \times 512$.

\begin{table}[ht]
\renewcommand{\arraystretch}{1.2}
\centering
\caption{{\em Allen-Cahn equation:} Hyper-parameter configuration for reproducing results in Figure \ref{fig: ac}.}
\label{tab: ac_config}
\begin{tabular}{ll}
\toprule
\textbf{Parameter} & \textbf{Value} \\
\midrule
\textbf{Architecture} & \\
Number of layers & 9 \\
Number of channels & 256 \\
Activation & Tanh \\
Fourier feature scale & 2.0 \\
Random weight factorization & $\mu =1.0, \sigma=0.1$ \\
\addlinespace  

\textbf{Learning rate schedule} & \\
Initial learning rate & $10^{-3}$ \\
Decay rate & 0.9 \\
Decay steps &  $5 \times 10^3$ \\
Warmup steps &   $5 \times 10^3$ \\
\addlinespace  

\textbf{Training} & \\
Training steps & $3 \times 10^5$ \\
Batch size & 8,192 \\
\addlinespace  

\textbf{Weighting} & \\
Weighting scheme & NTK \cite{wang2022and,wang2023expert} \\
Causal tolerance & 1.0 \\
Number of chunks & 32 \\

\bottomrule
\end{tabular}
\end{table}

\section{Korteweg–De Vries equation}

\paragraph{Data generation:}  We solve the Korteweg–De Vries equation using conventional spectral methods. Specifically, assuming periodic boundary conditions, we start from the initial condition $u_0(x) = \cos(\pi x)$ and integrate the system up to the final time $T=1$. Synthetic validation data is generated using the Chebfun package \cite{driscoll2014chebfun} with a spectral Fourier discretization with 512 modes and a fourth-order stiff time-stepping scheme (ETDRK4) \cite{cox2002exponential} with the time-step size of $10^{-5}$. We record the solution at intervals of 
 $\Delta t = 0.005$, yielding a validation dataset with a resolution of  $200 \times 512$ in temporal spatial domain..

\begin{table}[ht]
\renewcommand{\arraystretch}{1.2}
\centering
\caption{{\em Korteweg–De Vries equation:} Hyper-parameter configuration  for reproducing results in Figure \ref{fig:kdv}.}
\label{tab: kdv_config}
\begin{tabular}{ll}
\toprule
\textbf{Parameter} & \textbf{Value} \\
\midrule
\textbf{Architecture} & \\
Number of layers & 9 \\
Number of channels & 256 \\
Activation & Tanh \\
Fourier feature scale & 1.0 \\
Random weight factorization & $\mu =1.0, \sigma=0.1$ \\
\addlinespace  

\textbf{Learning rate schedule} & \\
Initial learning rate & $10^{-3}$ \\
Decay rate & 0.9 \\
Decay steps &  $2 \times 10^3$ \\
Warmup steps &   $5 \times 10^3$ \\
\addlinespace  

\textbf{Training} & \\
Training steps & $2 \times 10^5$ \\
Batch size & 4,096 \\
\addlinespace  

\textbf{Weighting} & \\
Weighting scheme & Grad norm \cite{wang2021understanding} \\
Causal tolerance & 1.0 \\
Number of chunks & 16 \\

\bottomrule
\end{tabular}
\end{table}

\section{Grey-Scott equation}

\paragraph{Data generation:}  We solve the Grey-Scott equation using conventional spectral methods. Specifically, assuming periodic boundary conditions, we start from the initial condition and integrate the system up to the final time $T=2$. Synthetic validation data is generated using the Chebfun package \cite{driscoll2014chebfun} with a spectral Fourier discretization with $200 \times 200$ modes in 2D and a fourth-order stiff time-stepping scheme (ETDRK4) \cite{cox2002exponential} with the time-step size of $10^{-3}$. We record the solution at intervals of 
 $\Delta t = 0.02$, yielding a validation dataset with a resolution of  $100 \times 200 \times 200$ in temporal spatial domain.

 \begin{table}[ht]
\renewcommand{\arraystretch}{1.2}
\centering
\caption{{\em  Grey-Scott equation:} Hyper-parameter configuration for reproducing results in Figure  \ref{fig:gs_pred}.}
\label{tab: gs_config}
\begin{tabular}{ll}
\toprule
\textbf{Parameter} & \textbf{Value} \\
\midrule
\textbf{Architecture} & \\
Number of layers & 9 \\
Number of channels & 256 \\
Activation & Swish \\
Fourier feature scale & 1.0 \\
Random weight factorization & $\mu =0.5, \sigma=0.1$ \\
\addlinespace  

\textbf{Learning rate schedule} & \\
Initial learning rate & $10^{-3}$ \\
Decay rate & 0.9 \\
Decay steps &  $2 \times 10^3$ \\
Warmup steps &   $5 \times 10^3$ \\
\addlinespace  

\textbf{Training} & \\
Number of time windows & $10$ \\ 
Training steps per time window & $ 10^5$ \\
Batch size & 4,096 \\

\addlinespace  

\textbf{Weighting} & \\
Scheme & Grad norm \cite{wang2021understanding} \\
Causal tolerance & 1.0 \\
Number of chunks & 32 \\

\bottomrule
\end{tabular}
\end{table}

\section{Ginzburg-Laudau equation}

\paragraph{Data generation:}  We solve the Ginzburg-Laudau equation using conventional spectral methods. Specifically, assuming periodic boundary conditions, we start from the initial condition and integrate the system up to the final time $T=1$. Synthetic validation data is generated using the Chebfun package \cite{driscoll2014chebfun} with a spectral Fourier discretization with $200 \times 200$ modes in 2D and a fourth-order stiff time-stepping scheme (ETDRK4) \cite{cox2002exponential} with time-step size of $10^{-3}$. We record the solution at intervals of 
 $\Delta t = 0.01$, yielding a validation dataset with a resolution of  $100 \times 200 \times 200$ in temporal spatial domain..

 \begin{table}[ht]
\renewcommand{\arraystretch}{1.2}
\centering
\caption{{\em  Ginzburg-Laudau equation:} Hyper-parameter configuration  for reproducing results in Figure  \ref{fig:gl_pred}.}
\label{tab: gl_config}
\begin{tabular}{ll}
\toprule
\textbf{Parameter} & \textbf{Value} \\
\midrule
\textbf{Architecture} & \\
Number of layers & 9 \\
Number of channels & 256 \\
Activation & Swish \\
Fourier feature scale & 1.0 \\
Random weight factorization & $\mu =0.5, \sigma=0.1$ \\
\addlinespace  

\textbf{Learning rate schedule} & \\
Initial learning rate & $10^{-3}$ \\
Decay rate & 0.9 \\
Decay steps &  $2 \times 10^3$ \\
Warmup steps &   $5 \times 10^3$ \\
\addlinespace  

\textbf{Training} & \\
Number of time windows & $ 5$ \\ 
Training steps per time window & $ 10^5$ \\
Batch size & 8,192 \\

\addlinespace  

\textbf{Weighting} & \\
Scheme & Grad norm \cite{wang2021understanding} \\
Causal tolerance & 5.0 \\
Number of chunks & 16 \\

\bottomrule
\end{tabular}
\end{table}

\section{Lid-driven Cavity flow}

\paragraph{Data generation:}  We compare our results against \cite{GHIA1982387}.

 \begin{table}[ht]
\renewcommand{\arraystretch}{1.2}
\centering
\caption{{\em  Lid-driven Cavity flow:} Hyper-parameter configuration  for reproducing results in Figure  \ref{fig:ldc_pred}.}
\label{tab: ldc_config}
\begin{tabular}{ll}
\toprule
\textbf{Parameter} & \textbf{Value} \\
\midrule
\textbf{Architecture} & \\
Number of layers & 18 \\
Number of channels & 256 \\
Activation & Tanh \\
Fourier feature scale & 15.0 \\
Random weight factorization & $\mu =1.0, \sigma=0.1$ \\
\addlinespace  

\textbf{Learning rate schedule} & \\
Initial learning rate & $10^{-3}$ \\
Decay rate & 0.9 \\
Decay steps &  $ 10^4$ \\
Warmup steps &   $5 \times 10^3$ \\
\addlinespace  

\textbf{Training} & \\
$Re$ & [100, 400, 1,000, 1,600, 3,200] \\
Training steps & [$ 10^4$, $2 \times10^4$, $5 \times10^4$, $5 \times10^4$, $5 \times 10^5$] \\
Batch size & 4,096 \\

\addlinespace  

\textbf{Weighting} & \\
Scheme & Grad norm \cite{wang2021understanding} \\

\bottomrule
\end{tabular}
\end{table}

\end{document}